\documentclass[review]{elsarticle}

\usepackage{lineno,hyperref}
\modulolinenumbers[5]

\journal{Journal of Image and Vision Computing}

\usepackage{graphicx}
\usepackage{amsmath,amssymb} 
\usepackage{color}

\usepackage{times}
\usepackage{epsfig}

\usepackage{algorithm}
\usepackage{subfiles}
\usepackage{nth}
\usepackage[noend]{algpseudocode}

\newcommand{\etal}{\mbox{\emph{et al.\ }}}

\begin{document}

\begin{frontmatter}

\title{Fine-Grained Categorization via CNN-Based Automatic Extraction and Integration of Object-Level and Part-Level Features}

\author{Ting Sun, Lin Sun, Dit-Yan Yeung}
\address{Hong Kong University of Science and Technology, Clear Water Bay, Hong Kong}




\begin{abstract}
Fine-grained categorization can benefit from part-based features which reveal subtle visual differences between object categories.  Handcrafted features have been widely used for part detection and classification.  Although a recent trend seeks to learn such features automatically using powerful deep learning models such as convolutional neural networks (CNN), their training and possibly also testing require manually provided annotations which are costly to obtain.  To relax these requirements, we assume in this study a general problem setting in which the raw images are only provided with object-level class labels for model training with no other side information needed.  Specifically, by extracting and interpreting the hierarchical hidden layer features learned by a CNN, we propose an elaborate CNN-based system for fine-grained categorization.  When evaluated on the Caltech-UCSD Birds-200-2011, FGVC-Aircraft, Cars and Stanford dogs datasets under the setting that only object-level class labels are used for training and no other annotations are available for both training and testing, our method achieves impressive performance that is superior or comparable to the state of the art.  Moreover, it sheds some light on ingenious use of the hierarchical features learned by CNN which has wide applicability well beyond the current fine-grained categorization task.
\end{abstract}


\end{frontmatter}

\linenumbers

\section{Introduction}
\label{sec:Introduction}
Fine-grained visual categorization refers to a special type of image classification tasks in which the object categories generally have the same constituent parts and topology and hence they are visually and semantically very similar to each other. Some examples include fine distinction of birds according to species and human faces according to age or gender.  Because different categories are very similar, fine-grained categorization often requires identifying subtle part-based differences between categories. This is particularly challenging when there exist large within-category variations such as pose and scale variance, or when only a small amount of training data is available.

Consequently, many fine-grained categorization systems put their emphasis on exploiting part-based features to boost the classification performance. One common approach is to make use of manually provided annotations, such as a bounding box around the whole object and the part locations indicated by coordinates, to extract fine features from the object parts \cite{Birdlet,POOF,symbiotic,birdspecies,decaf,nonparametric,lin2015deep,zhang2016spda,yao2016coarse,huang2016part,huang2016pbc}.  The main limitation of this approach is that it is laborious to obtain data with part annotations to provide part-based features for the classifier.  Moreover, for images outside the dataset with no part annotations available, applying the trained classifier on them requires using a part detector first. Although having such a part detector makes it unnecessary to provide part annotations for the test images, training the part detector still needs images with part annotations available.  Some recent attempts have been made to relax this requirement to use less annotations by training a part detector in a weakly supervised manner \cite{sermanet2014attention,krause2014learning,krause2015fine,simon2015neural,wang2015multiple,jaderberg2015spatial,weakly,wang2016mining,luo2016annotation,huang2016task,zhang2016picking,zhang2016fused,xu2017friend}. As a result, part locations are no longer needed for both training and testing.  However, inaccurate part locations returned by the part detector can affect the quality of the part-based features extracted and hence impair the classification performance.  To reduce this effect, computationally demanding procedures are often integrated into the system to improve the part detection accuracy. Without part annotations, a typical way to identify the object parts is to group or select them from randomly or exhaustively generated part proposals. It is costly to verify massive part proposals without much guidance. It would be beneficial if we could develop a more informed process for part detection by exploiting the features automatically learned solely from the raw images and the corresponding category labels. This leads us to a more promising alternative which aims at developing an automatic yet efficient process for part detection with minimal supervision required.

Deep learning \cite{lecun2015deep,schmidhuber2015deep} has an important role to play here because it focuses on learning features or representations directly from raw data and it learns the features in a hierarchical manner supervised by the category labels only. In particular, convolutional neural networks (CNN or ConvNet) \cite{le1990handwritten} provide a powerful end-to-end framework which tightly integrates feature extraction and classification to achieve state-of-the-art performance in many challenging computer vision tasks \cite{AlexNet,decaf,razavian2014cnn,VGG,GoogLeNet,jaderberg2015spatial}.  It should be noted that the excitement does not just come from the superior performance achieved.  The rich features learned by deep CNN ranging from low-level to high-level representations in the hidden layers have also aroused much research interest in investigating how to take advantage of them \cite{zeiler2014visualizing,weakly,zhou2014object,wang2015multiple,cimpoi2015deep,wang2015action,xu2015discriminative,guo2016locally,zhou2016learning}.  On one hand, interpreting the hidden layer features may help us understand the workings of CNNs and monitor the learning process.  On the other hand, exploiting the features appropriately may help to further boost the performance of various tasks.  However, due to the complexity of the highly encapsulated CNN architecture especially when it is deep, exploiting the hidden layer features learned turns out to be highly nontrivial. The recent attempts either show some general results without concrete methods for specific tasks  \cite{deepinside,decaf,simon2014part} or exhaustively use the features learned by a CNN to come up with region proposals to assist some other manually designed tools \cite{birdspecies,razavian2014cnn,zhang2014part,krause2015fine,simon2015neural,huang2016task,zhang2016picking}.

We believe a well-trained CNN has potential that remains to be more fully explored and this has motivated us to pursue the current research.  Our contribution in this work is twofold.  First we explore the hidden layer feature maps of a well-trained CNN more thoroughly and make good use of them in simple but nontrivial ways.  We visualize not only the hidden layer feature maps as they are to conclude their properties, but also the intermediate results of each step when we manipulate them.  Second, guided by the general observation of CNN, we propose a pure CNN-based system tailed for fine-grained categorization.  Partially, our system obtains robust object and parts detection through \textit{interpreting} the hidden layers of a trained CNN rather then \textit{calculating} them, which often requires a lot more computation, parameters and annotations.

The remainder of this paper is organized as follows.  Sec.~\ref{sec:Related work} reviews some previous work on fine-grained categorization and the related CNN study. Our proposed method is presented in Sec.~\ref{sec:Proposed system} which is then followed by experiment results in Sec.~\ref{sec:Experiments}. Sec.~\ref{sec:Conclusion} concludes the paper. 

\section{Related Work}
\label{sec:Related work}
The approaches to handle fine-grained categorization vary a lot while the essentialness of part-based feature is well recognized.   Various handcrafted features have been involved in related part-based methods \cite{Birdlet,codebook,zhang2012pose,POOF,iscen2015comparison}.  For example, part-based one-vs-one features (POOF) \cite{POOF,crow,Birdsnap} are based on histograms of oriented gradients (HOG) \cite{2005histograms,2010object} and color histograms.  Each POOF is defined based on the locations of two parts which are used to align the object and a POOF is extracted around one of the two parts.  The success of POOF features demonstrates the effectiveness of part-based features for fine-grained categorization, but its need for manual selection and location of the parts makes it somewhat unappealing.

After CNN was demonstrated to give superior performance in the ImageNet challenge \cite{AlexNet}, we have witnessed a resurgence of research interest in CNN \cite{deepinside,decaf,zeiler2014visualizing,liu2015deep}.  Moreover, the public availability of efficient CNN implementations \cite{caffe} and powerful pre-trained CNN models \cite{AlexNet,chatfield2014return,VGG,GoogLeNet} has further popularized the pervasive use of CNN for various computer vision tasks.  Not surprisingly, fine-grained categorization also turns to CNN for automatic feature extraction and part detection.  Branson \etal~\cite{birdspecies} use CNN to extract local features from pose-normalized images.  An object is first warped to a pose prototype where both the warping function and the pose prototype are learned by minimizing some manually designed objective functions with the part locations needed.  Based only on segmentation, Krause \etal~\cite{krause2015fine} locate the parts through alignment. No part annotations are needed for both training and testing, but complex co-segmentation, alignment and part selection methods are involved and the ground-truth object bounding box is crucial to proper initialization and refinement of the segmentation procedure. Simon \etal~\cite{simon2014part,simon2015neural} use the hidden layer feature maps of a CNN for part detection.  Assuming that some feature maps of the CNN correspond to object parts, they use the gradient map of each feature map to offer a part proposal.  A likelihood function is used to select the maps that correspond to valid object parts.  However, although a filter (corresponding to a feature map) can detect a certain pattern in an object part, there often does not exist a simple one-to-one correspondence between the feature maps and the object parts.  Thus, selecting one map to locate one object part may not be sufficient. Sermanet \etal~\cite{overfeat} use a CNN to automatically find a bounding box for the object by combining the results obtained from many sliding windows. Unlike selecting only one choice in \cite{simon2014part,simon2015neural}, combining multiple feature maps provides a more robust approach which is also adopted by our method.

Some recent methods achieve promising performance without requiring additional information other than the object labels
\cite{xiao2014application,sermanet2014attention,bilinear,simon2015neural,weakly,wang2015multiple}. The motivation of the bilinear CNN model \cite{bilinear} is to use two CNNs to factor out the variance due to location and the appearance of the object parts.  However, it is mentioned in \cite{bilinear} that the roles of the two networks are not clearly separated.
A CNN is used to detect both the whole object and its parts in \cite{weakly} and \cite{xiao2014application}. The initial object proposals in \cite{weakly}, which include many noisy background regions, are generated by selective search \cite{selectivesearch}. The two-level attention model \cite{xiao2014application} is based on the intuition that performing fine-grained categorization requires first to ``see'' the object and then the most discriminative parts of it.  This intuition is also crucial in our system design. In \cite{xiao2014application} the feature maps are clustered according to their filter coefficients, but the same object part may not have similar texture for different classes.

Although CNN has been used recently for fine-grained categorization, we believe there is still much room for further investigation especially on using CNN for automatic part detection.  Our preliminary investigation shows that the hidden layer features extracted from a raw image by a well-trained CNN are rich but implicit. Unlike the classifier, the hidden layers are not trained to output an explicit target.  Consequently, using them for part detection requires a carefully designed procedure to interpret the hidden layer features learned by a CNN.  The filters respond to the corresponding patterns that appear in their field of perception, but those patterns may reside in both the foreground and the background and they may only correspond to partial regions inside object parts.  In the previous work, a simplifying assumption is adopted by using one feature map as one part proposal and all the proposals are verified extensively. Instead, we take advantage of the hierarchical features extracted by a CNN based on the object detection result to impose constraints on part detection and sum over all the selected feature maps to achieve robust part detection.

\section{Proposed System}
\label{sec:Proposed system}
\begin{figure}[h!!]
 \centering
 \includegraphics[width=0.8\textwidth]{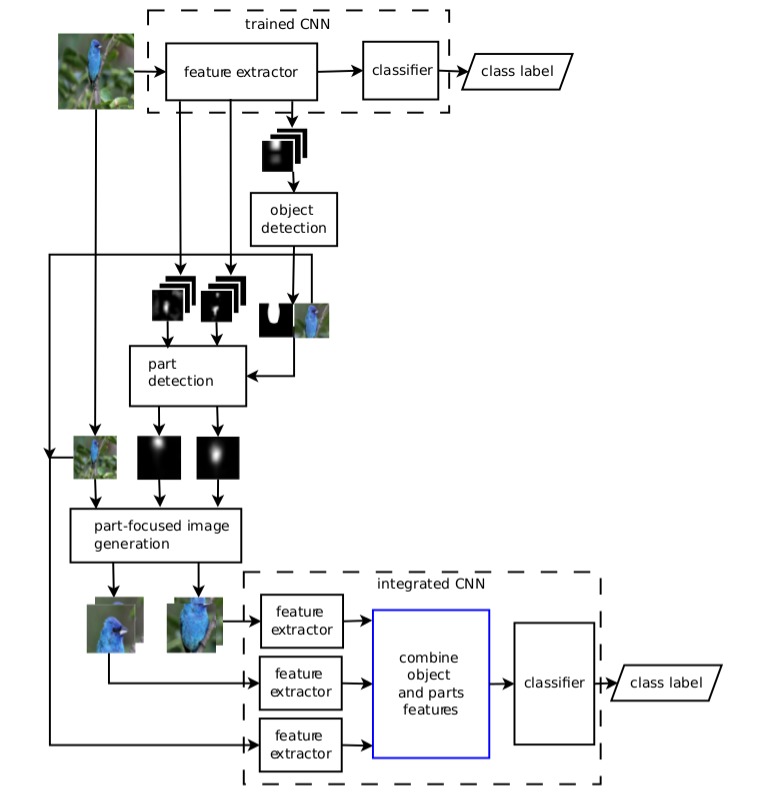}
 \caption{Schematic diagram shows the proposed system}
 \label{fig:sys}
\end{figure}

Our fine-grained categorization system is entirely based on CNNs trained using raw images and their class labels only. A CNN is first trained using raw images as input. The hidden layer feature maps of the trained CNN are then used to detect the object and its parts. Based on the detection result, we crop some image patches each of which tightly contains either the whole object or one object part. These generated images are referred to as object-focused and part-focused images. As such, the original images are augmented by focusing on the targets at different levels.  We then construct an additional CNN for each object part detected and initialize it with the CNN previously trained using raw images, and then fine-tune it with the corresponding part-focused images.  All the CNNs are combined into an integrated CNN which extracts and combines both object-level and part-level features before classification.  The whole system is shown in Figure~\ref{fig:sys} with two parts detected.

In the following subsections, we will first present clearly the notation and the initialization setting of our algorithm.  We will then highlight the major steps and the key ideas behind them.

\subsection{Notation and System Initialization}
When describing our algorithm below, the image, mask, index set, hidden layer feature map, and 2-D centroid coordinates (in an image) are denoted by $\mathbf{I}$, $\mathbf{M}$, $\mathcal{S}$, $\mathbf{F}$, and $\mathbf{c}$, respectively.
Superscripts and subscripts are used to identify different variables of the same kind and also give information about their properties. $\mathbf{I}_{orig}$ represents the original image.  Its $k$th part-focused image generated is $\mathbf{I}_{part\_k}$. $\mathcal{S}^{layer\_idx}_{object}$ specifies the set of indices of the layers from whose features the whole object is detected while $\mathcal{S}^{layer\_idx}_{parts}$ specifies those used to detect the parts. Figure~\ref{fig:sys} shows the case with $|\mathcal{S}^{layer\_idx}_{object}|=1$ and $|\mathcal{S}^{layer\_idx}_{parts}|=2$, where $|\cdot|$ denotes the cardinality of a set. $\mathbf{F}^j_i$ means the $i$th feature map in layer $j$ and $\mathbf{c}_{j,i}$ is its weighted centroid. In addition to the given input, $\mathcal{S}^{map\_idx}_{layer\_j}$ contains all the feature map indices in layer $j$. All the other sets like $\mathcal{S}^{map\_idx}_{valid}$ and $\mathcal{S}^{centroid}_{valid}$ are used to store the intermediate results and are initialized as empty sets. $N_{part}$ denotes the number of parts to detect. $T_{object}$ and $T_{parts}$ are two scalar thresholds. $\mathcal{N}(\cdot)$ is a normalization function which scales each entry of its input matrix (representing a grayscale image) to the range $[0,1]$, and $\mathcal{R}(\cdot)$ is a function which resizes its input (both color and grayscale images) so that the images have the same resolution as the CNN input.

\subsection{Object Detection}
An active region in a feature map indicates the existence of a certain pattern detected at that location. A well-trained CNN should make good use of the patterns that are useful for distinguishing between different categories. Since the background is generally irrelevant to object categorization, the active regions usually reside within the object. Although some background regions may contain patterns similar to those in the object, the background patterns usually correspond to low-level features and the corresponding regions are unlikely to remain active in the higher-level feature maps. This phenomenon will be illustrated in Subsec~\ref{subsubsec: set choice}. The object detection result is given by a saliency mask, which is essentially a $[0,1]$ grayscale image resized to the same resolution as the CNN input.  The brighter the region in a mask, the stronger the indication of existence of its target.  In our method, the object saliency mask $\mathbf{M}_{object}$ and its binary version $\mathbf{M}^{binary}_{object}$ are calculated based on steps (\ref{equ: obj layer sum}) through (\ref{equ: obj mask binary}) below:

\begin{equation}
 \label{equ: obj layer sum}
 \mathbf{M}_{layer, sum}^{j} = \sum\limits_{i\in \mathcal{S}^{map\_idx}_{layer\_j}} \mathbf{F}^j_i
\end{equation}
\begin{equation}
\label{equ: obj layer sum normalize}
 \mathbf{M}_{layer, sum}^{j} =  \mathcal{N}(\mathbf{M}_{layer, sum}^{j})
\end{equation}
\begin{equation}
\label{equ: obj layer sum resize}
 \mathbf{M}_{layer, sum}^{j} =  \mathcal{R}(\mathbf{M}_{layer, sum}^{j})
\end{equation}
\begin{equation}
\label{equ: obj mask calculate}
 \mathbf{M}_{object} = \prod\limits_{j\in \mathcal{S}^{layer\_idx}_{object}} \mathbf{M}_{layer, sum}^{j}
\end{equation}
\begin{equation}
\label{equ: obj mask normalize}
 \mathbf{M}_{object} = \mathcal{N}(\mathbf{M}_{object})
\end{equation}
\begin{equation}
\label{equ: obj mask binary}
 \mathbf{M}^{binary}_{object}(x,y) =
   \begin{cases}
     1, & \text{if }\ \mathbf{M}_{object}(x,y) >  T_{object}\\
     0, & \text{otherwise.} 
   \end{cases}
\end{equation}

For each layer $j\in \mathcal{S}^{layer\_idx}_{object}$ (How we choose $\mathcal{S}^{layer\_idx}_{object}$ and other parameters are detailed in Subsec~\ref{subsec: parameter choice}), all its feature maps are summed and then normalized to obtain $\mathbf{M}_{layer, sum}^{j}$.  In so doing, all the patterns are aggregated and the activations in the background are suppressed after normalization since they are overshadowed by those within the object.  In (\ref{equ: obj mask calculate}), the notation $ \prod\limits_{j\in \mathcal{S}^{layer\_idx}_{object}} $ means applying element-wise multiplication to all the summed maps in all the layers indexed by $\mathcal{S}^{layer\_idx}_{object}$ to further increase the confidence. Consequently, an active region in the resulting mask $\mathbf{M}_{object}$ has to be active in all the layers in $\mathcal{S}^{layer\_idx}_{object}$.  

\subsection{Part Detection}
\label{subsec:Part detection}
The key component of our system is to automatically detect the object parts using the hidden layer features of the CNN trained with raw images and their class labels only. We first choose the feature maps which are likely to be activated by the object parts and then cluster them into $N_{part}$ groups. The mask of the $k$th part $\mathbf{M}_{part\_k}$ is obtained by summing over all the feature maps within the corresponding cluster.  The detailed steps involved in part detection are listed in Algorithm~\ref{alg:mask generation}.

\begin{algorithm}[h!!]
\caption{Part detection}
\label{alg:mask generation}
\begin{algorithmic}[1]
\Require ~~\ 
$\mathbf{F}^j_i$, $\mathcal{S}^{layer\_idx}_{parts}$, $N_{part}$, $T_{parts}$, $\mathbf{M}^{binary}_{object}$
\Ensure ~~\ 
$\mathbf{M}_{part\_k}$, $\mathbf{M}^{binary}_{part\_k}$. $(k = 1,2,...,N_{part})$ 
\ForAll {$j \in \mathcal{S}^{layer\_idx}_{parts}, i \in \mathcal{S}^{map\_idx}_{layer\_j}$} 
\State $\mathbf{F}^j_i =  \mathcal{N}(\mathbf{F}^j_i)$ 
\State $\mathbf{F}^j_i =  \mathcal{R}(\mathbf{F}^j_i)$
\State \begin{equation}
         \mathbf{F}^{j, binary}_i(x,y) = 
         \begin{cases}
          1, & \text{if }\  \mathbf{F}^j_i(x,y) > T_{parts}\\
          0, & \text{otherwise.}
         \end{cases}   \nonumber
       \end{equation}
\State $\mathbf{c}_{j,i} \gets  \textit{weighted centroid of } \mathbf{F}^j_i \textit{ in the positive region of } \mathbf{F}^{j, binary}_i$
\If {$\mathbf{F}^{j, binary}_i \textit{ contains one connected region }$}
\If {$\mathbf{c}_{j,i} \textit{ is in the positive region of } \mathbf{M}^{binary}_{object}$} 
\State $\mathcal{S}^{map\_idx}_{valid} = \mathcal{S}^{map\_idx}_{valid} \cup (j,i)$
\State $\mathcal{S}^{centroid}_{valid} = \mathcal{S}^{centroid}_{valid} \cup \mathbf{c}_{j,i}$
\EndIf
\EndIf
\EndFor
\State \Comment{Apply $k$-means to the elements in $\mathcal{S}^{centroid}_{valid}$}
\State $\mathcal{S}^{centroid}_{valid, part\_k} \xleftarrow{\textit{apply k-means}} \mathcal{S}^{centroid}_{valid} $
\State where $k=1, 2, ..., N_{part}$
\State $\mathcal{S}^{map\_idx}_{valid, part\_k} \xleftarrow{\textit{cluster the maps the same as corresponding centroid clusters}} \mathcal{S}^{map\_idx}_{valid} $
\State $\mathbf{M}_{part\_k} = \sum\limits_{(j, i)\in \mathcal{S}^{map\_idx}_{valid, part\_k}} \mathbf{F}^j_i $
\State $\mathbf{M}_{part\_k} = \mathcal{N}(\mathbf{M}_{part\_k})$
\State \begin{equation}
        \mathbf{M}^{binary}_{part\_k}(x,y) =
         \begin{cases}
          1, & \text{if }\ \mathbf{M}_{part\_k}(x,y) >  T_{parts}\\
          0, & \text{otherwise.} 
         \end{cases}  \nonumber
       \end{equation}
\State \textbf{return} $\mathbf{M}_{part\_k}, \mathbf{M}^{binary}_{part\_k}, k = 1,2,...,N_{part}$ 
\end{algorithmic}
\end{algorithm}

\begin{figure}[h!]
 \centering
 \includegraphics[width=0.062\textwidth, height=0.062\textwidth]{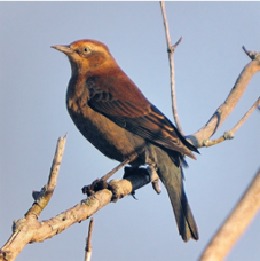}
 \includegraphics[width=0.062\textwidth, height=0.062\textwidth]{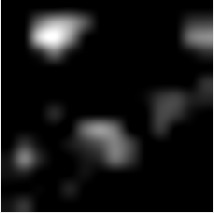}
 \includegraphics[width=0.062\textwidth, height=0.062\textwidth]{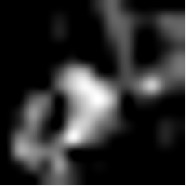} 
 \includegraphics[width=0.062\textwidth, height=0.062\textwidth]{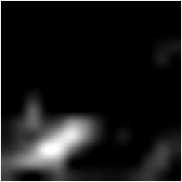}
 \includegraphics[width=0.062\textwidth, height=0.062\textwidth]{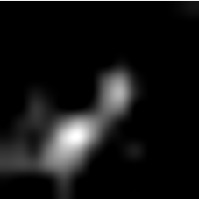}
 \includegraphics[width=0.062\textwidth, height=0.062\textwidth]{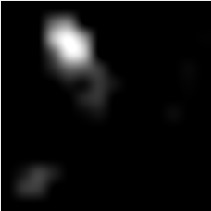} 
 \includegraphics[width=0.062\textwidth, height=0.062\textwidth]{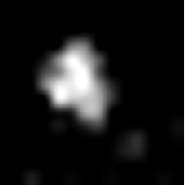} 
 \includegraphics[width=0.008\textwidth, height=0.062\textwidth]{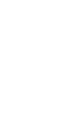} 
 \includegraphics[width=0.062\textwidth, height=0.062\textwidth]{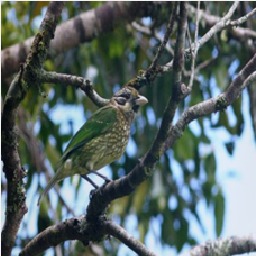}
 \includegraphics[width=0.062\textwidth, height=0.062\textwidth]{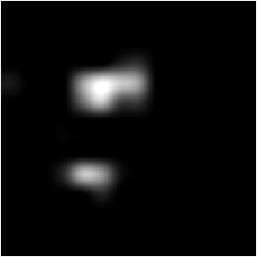}
 \includegraphics[width=0.062\textwidth, height=0.062\textwidth]{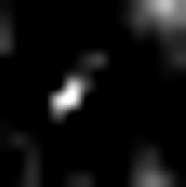} 
 \includegraphics[width=0.062\textwidth, height=0.062\textwidth]{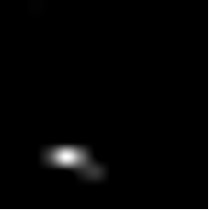}
 \includegraphics[width=0.062\textwidth, height=0.062\textwidth]{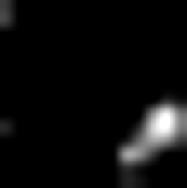}
 \includegraphics[width=0.062\textwidth, height=0.062\textwidth]{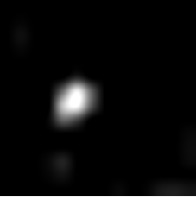} 
 \includegraphics[width=0.062\textwidth, height=0.062\textwidth]{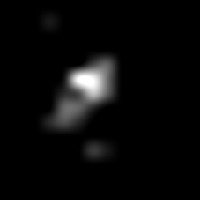} \\
 \includegraphics[width=0.062\textwidth, height=0.062\textwidth]{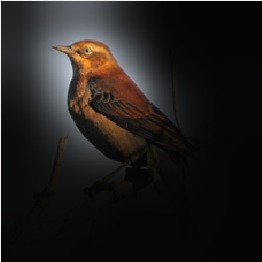}
 \includegraphics[width=0.062\textwidth, height=0.062\textwidth]{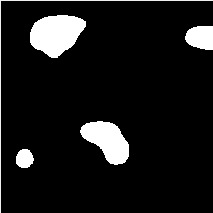}
 \includegraphics[width=0.062\textwidth, height=0.062\textwidth]{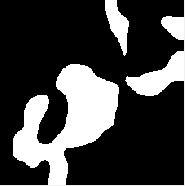} 
 \includegraphics[width=0.062\textwidth, height=0.062\textwidth]{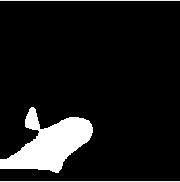}
 \includegraphics[width=0.062\textwidth, height=0.062\textwidth]{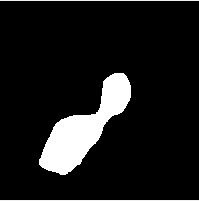}
 \includegraphics[width=0.062\textwidth, height=0.062\textwidth]{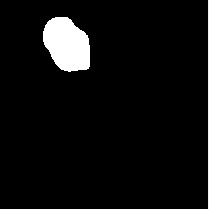} 
 \includegraphics[width=0.062\textwidth, height=0.062\textwidth]{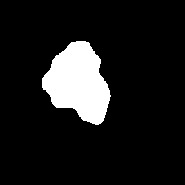} 
 \includegraphics[width=0.008\textwidth, height=0.062\textwidth]{white_space.jpg}
 \includegraphics[width=0.062\textwidth, height=0.062\textwidth]{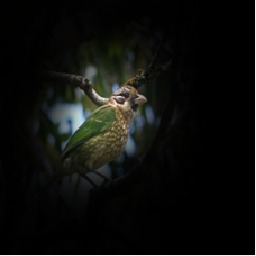}
 \includegraphics[width=0.062\textwidth, height=0.062\textwidth]{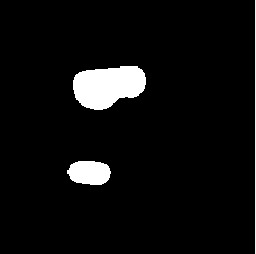}
 \includegraphics[width=0.062\textwidth, height=0.062\textwidth]{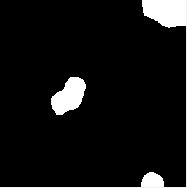} 
 \includegraphics[width=0.062\textwidth, height=0.062\textwidth]{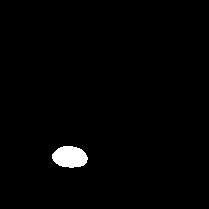}
 \includegraphics[width=0.062\textwidth, height=0.062\textwidth]{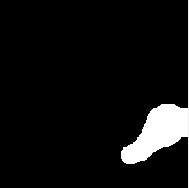}
 \includegraphics[width=0.062\textwidth, height=0.062\textwidth]{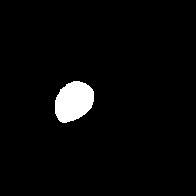} 
 \includegraphics[width=0.062\textwidth, height=0.062\textwidth]{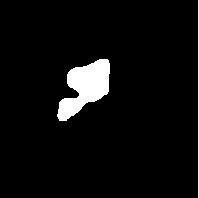} \\
 \includegraphics[width=0.062\textwidth, height=0.062\textwidth]{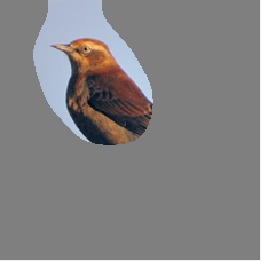}
 \includegraphics[width=0.062\textwidth, height=0.062\textwidth]{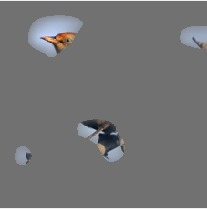}
 \includegraphics[width=0.062\textwidth, height=0.062\textwidth]{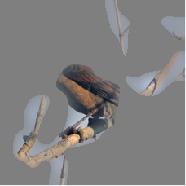} 
 \includegraphics[width=0.062\textwidth, height=0.062\textwidth]{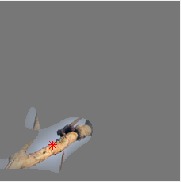}
 \includegraphics[width=0.062\textwidth, height=0.062\textwidth]{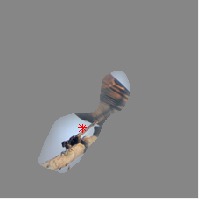}
 \includegraphics[width=0.062\textwidth, height=0.062\textwidth]{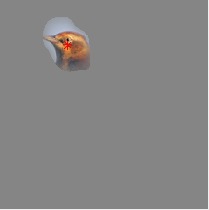} 
 \includegraphics[width=0.062\textwidth, height=0.062\textwidth]{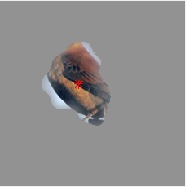} 
 \includegraphics[width=0.008\textwidth, height=0.062\textwidth]{white_space.jpg}
 \includegraphics[width=0.062\textwidth, height=0.062\textwidth]{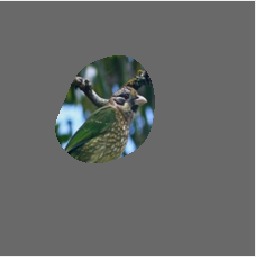}
 \includegraphics[width=0.062\textwidth, height=0.062\textwidth]{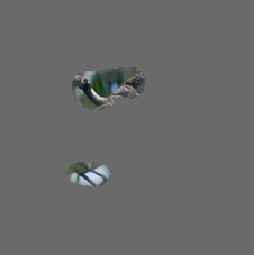}
 \includegraphics[width=0.062\textwidth, height=0.062\textwidth]{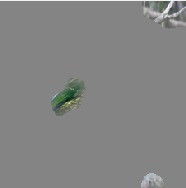} 
 \includegraphics[width=0.062\textwidth, height=0.062\textwidth]{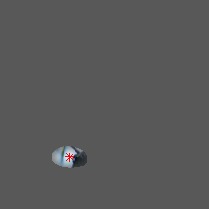}
 \includegraphics[width=0.062\textwidth, height=0.062\textwidth]{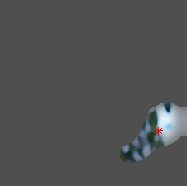}
 \includegraphics[width=0.062\textwidth, height=0.062\textwidth]{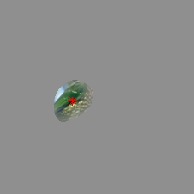} 
 \includegraphics[width=0.062\textwidth, height=0.062\textwidth]{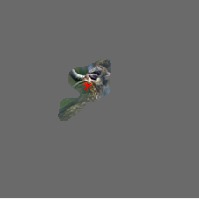} \\
 \caption{Illustration of feature map selection for part detection using two examples.  Detailed description can be found in the text.}
 \label{fig:sample masks}
\end{figure}

The layers used to detect object parts are from $\mathcal{S}^{layer\_idx}_{parts}$ (how to choose $\mathcal{S}^{layer\_idx}_{parts}$ is detailed in Subsec~\ref{subsubsec: set choice}), but not all their feature maps are useful for part detection because those with activations in the background may hurt the part detection accuracy during clustering. Thus we choose the maps that are likely to contribute to part detection based on two constraints.  First, the activated region in the thresholded map should be a connected region.  Second, this region should have its weighted centroid within the object detected previously. The selection results are stored in the set $\mathcal{S}^{map\_idx}_{valid}$.  Figure~\ref{fig:sample masks} illustrates this step using two examples, with each example containing $7$ columns. In the first column of each example, from top to bottom shows the original input image, the soft masked image, and the thresholded masked image given by object detection with $T_{object}=0.3$, respectively. (Our system is quite robust to the choice of $T_{object}$ and $T_{parts}$.  More details can be found in Sec.~\ref{sec:Experiments}.)
For column $2$ to $7$, the first row shows some normalized feature maps, the second row shows their thresholded version with $T_{parts}=0.3$, and the last row shows the results of applying the mask on the image with the mask's weighted centroid marked in red. Column~$2$ and $3$ show two maps that contain multiple disconnected active regions and thus fail to meet the first constraint.
Although these maps may be activated by the object parts, it is likely that they correspond to some patterns shared by multiple parts instead of just one, making them ineffective for part detection.  Column $4$ and $5$ show two feature maps that do contain only one connected region in their thresholded version, but their weighted centroids fail to fall within the detected object region. Notice that we only constrain the centroids but the actual active regions are allowed to extend beyond the detected object region.  This strategy increases the robustness and flexibility of our method. The last two columns show two feature maps which are successfully selected to detect object parts. In fact the hidden layer feature maps of the CNN are very rich. Even when some feature maps are ignored by mistakes, the aggregation result is still quite robust.

\begin{figure}[h!]
\centering
 \includegraphics[width=0.85\textwidth, height=0.1\textwidth]{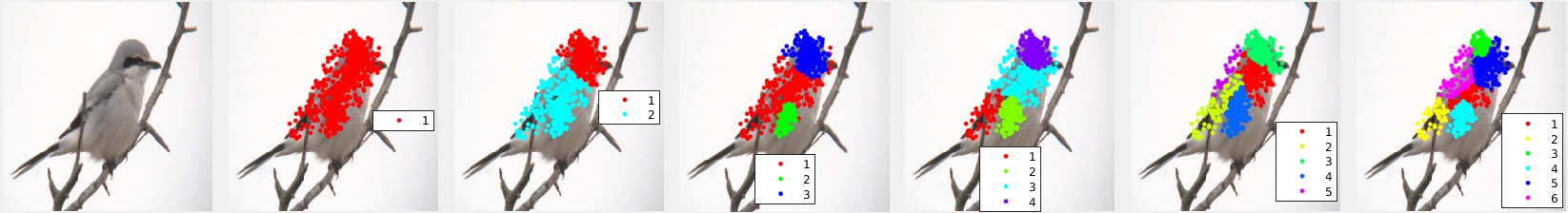}
 \includegraphics[width=0.85\textwidth, height=0.1\textwidth]{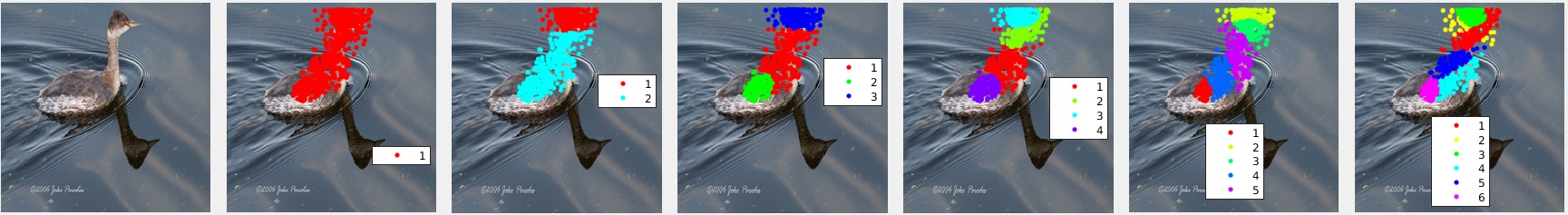}
 \includegraphics[width=0.85\textwidth, height=0.1\textwidth]{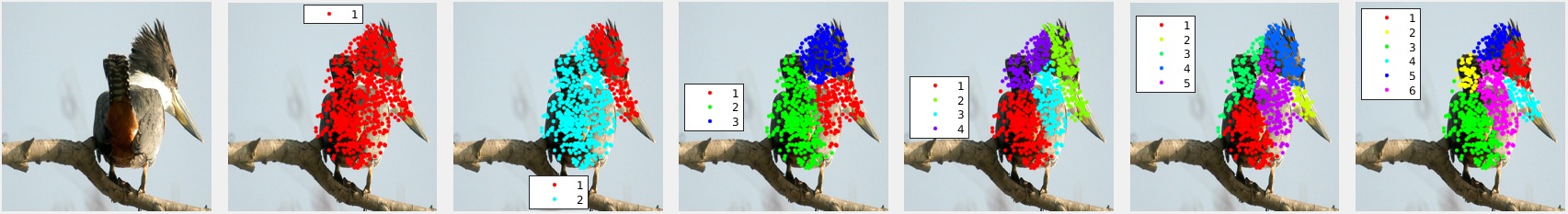}

 \includegraphics[width=0.85\textwidth, height=0.1\textwidth]{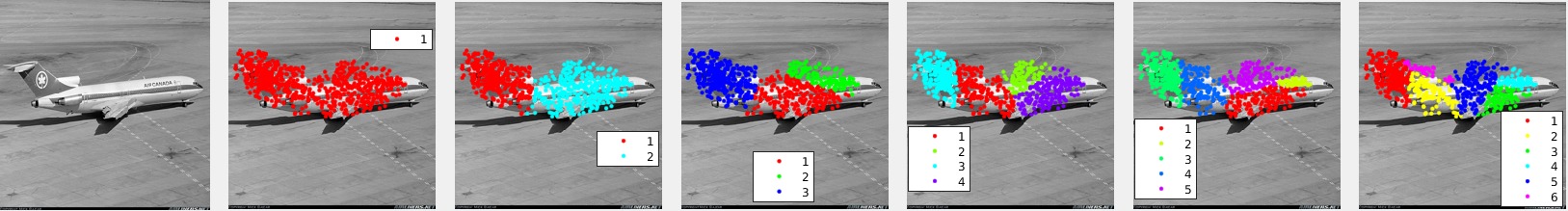}
 \includegraphics[width=0.85\textwidth, height=0.1\textwidth]{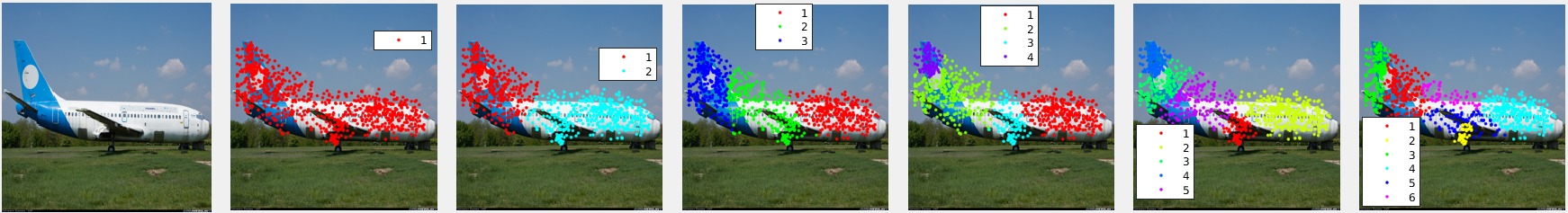}
 \includegraphics[width=0.85\textwidth, height=0.1\textwidth]{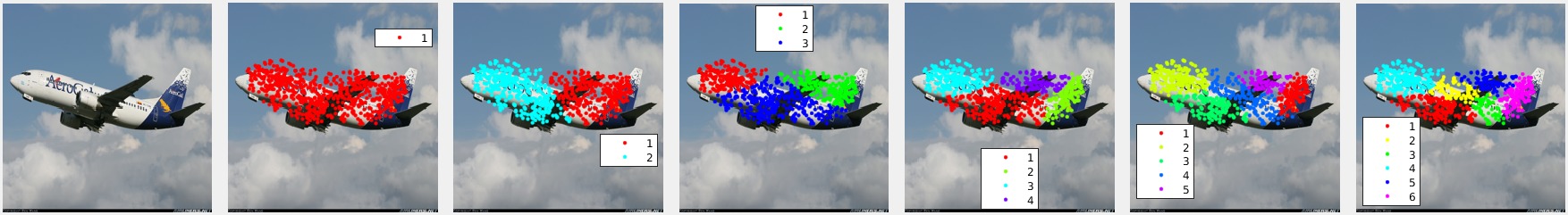}

 \caption{Clustering samples with $N_{part} = 1, 2, 3, 4, 5, 6$ of Caltech-UCSD Birds-200-2011 \cite{WelinderEtal2010} and FGVC-Aircraft \cite{fgvcaircraft} datasets.}
 \label{fig:clustering result}
\end{figure}

\begin{figure}[h!]
 \centering
 \includegraphics[width=0.8\textwidth, height=0.08\textwidth]{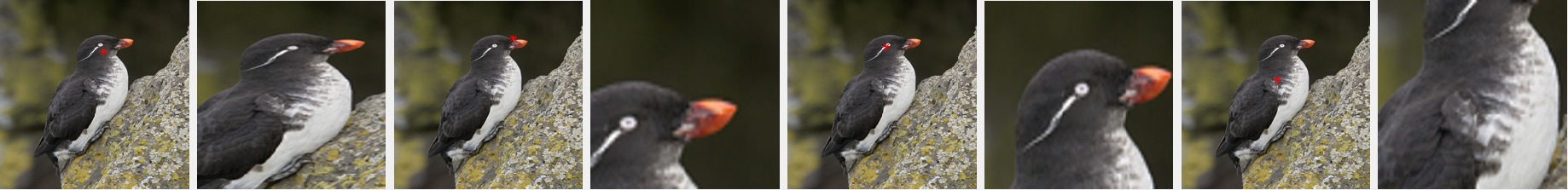}
 \includegraphics[width=0.8\textwidth, height=0.08\textwidth]{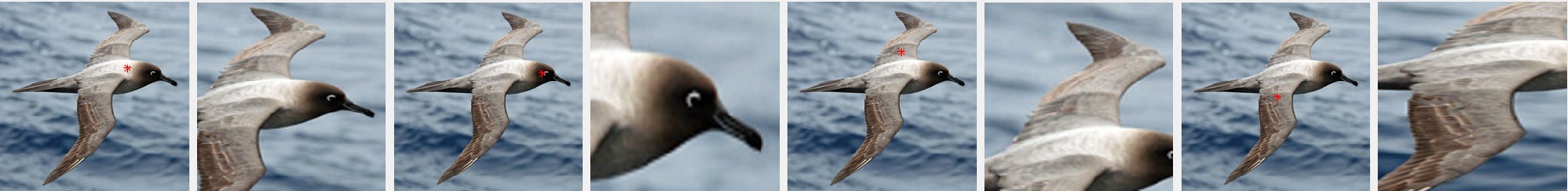}
 \includegraphics[width=0.8\textwidth, height=0.08\textwidth]{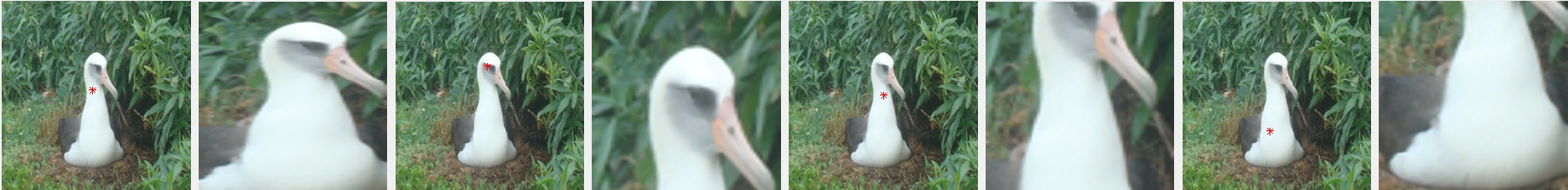}
 \includegraphics[width=0.8\textwidth, height=0.08\textwidth]{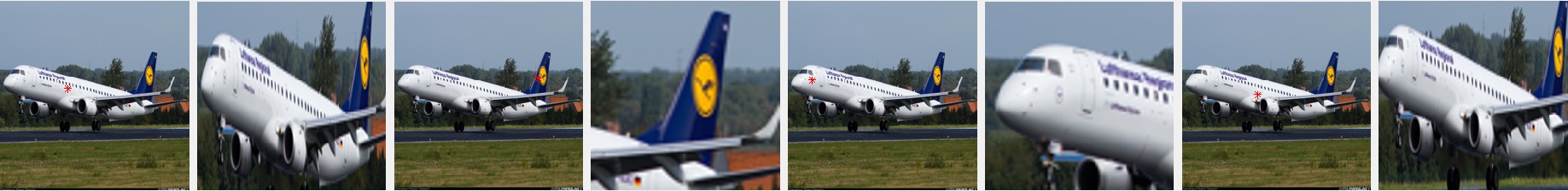}
 \includegraphics[width=0.8\textwidth, height=0.08\textwidth]{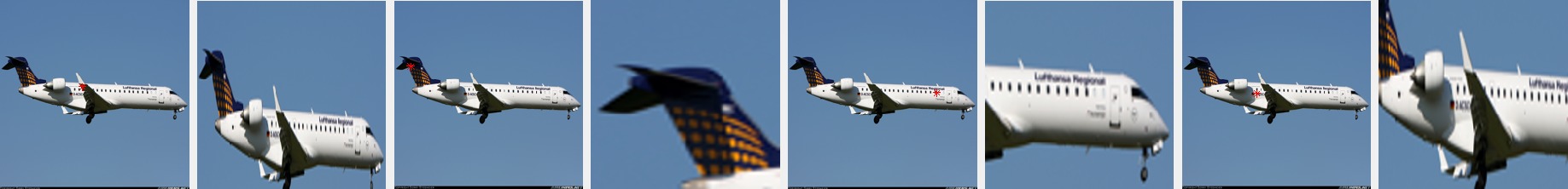}
 \includegraphics[width=0.8\textwidth, height=0.08\textwidth]{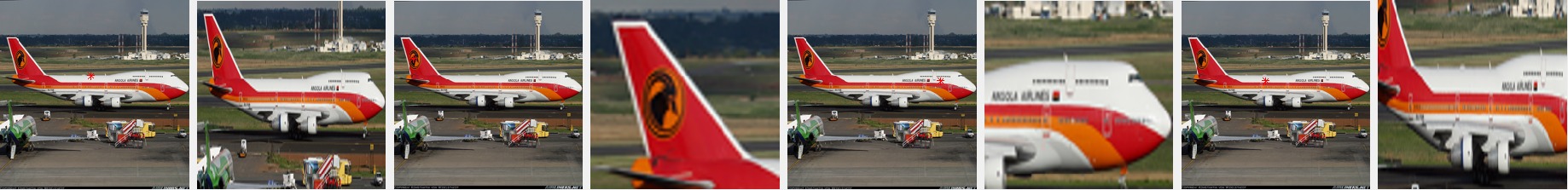}
 
 \includegraphics[width=0.8\textwidth, height=0.02\textwidth]{white_space.jpg}
 
 \includegraphics[width=\textwidth, height=0.08\textwidth]{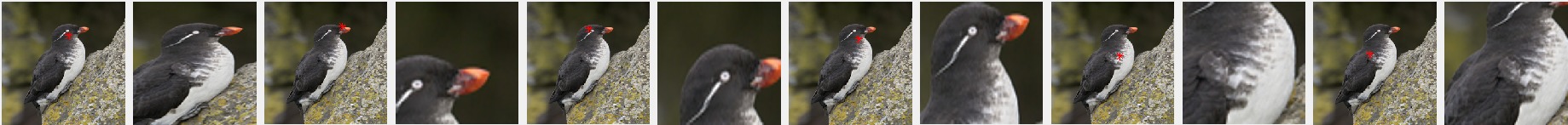}
 \includegraphics[width=\textwidth, height=0.08\textwidth]{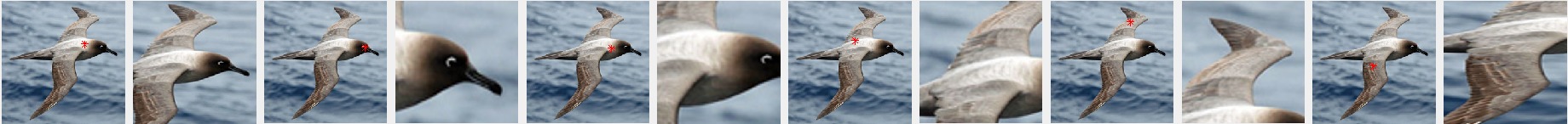}
 \includegraphics[width=\textwidth, height=0.08\textwidth]{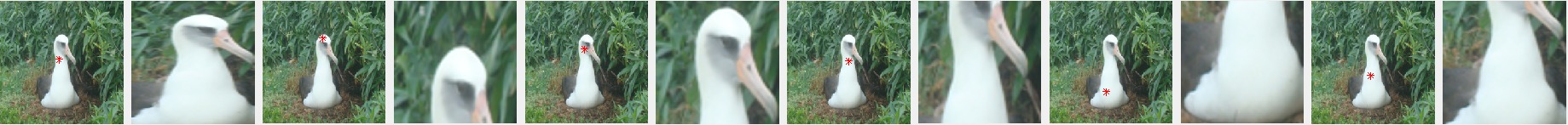}
 \includegraphics[width=\textwidth, height=0.08\textwidth]{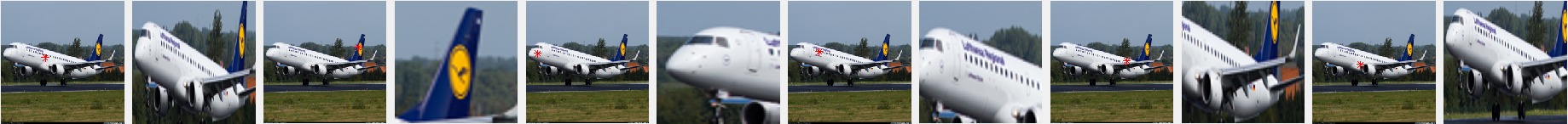}
 \includegraphics[width=\textwidth, height=0.08\textwidth]{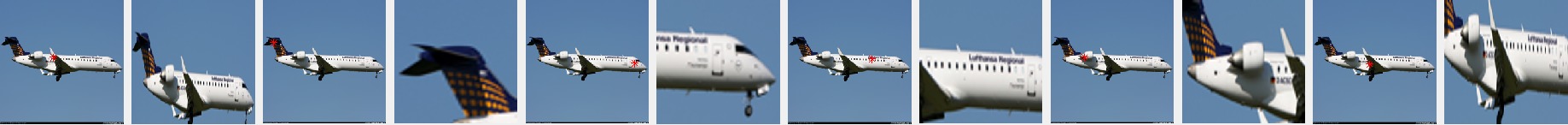}
 \includegraphics[width=\textwidth, height=0.08\textwidth]{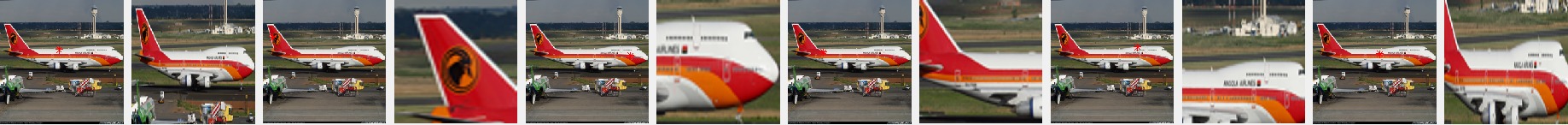}
 \caption{Generated object-focused and part-focused images with $N_{part} = 3, 5$. The top group of images shows the results of $N_{part} = 3$ and the down group shows that of $N_{part} = 5$. Each line of results corresponding to one example. The first and second images of each line are the original image and cropped object-focused image. The following images should be viewed in pairs, with the first one shows the part centroid in the original image and the second one is the generated part-focused image.}
 \label{fig:generated part-focused images}
\end{figure}

We cluster the selected feature maps' centroids by $k$-means so that the maps activated by the same part can be combined together to offer more robust detection of one part.  The clustering results of $N_{part} = 1, 2, 3, 4, 5, 6$ are shown in Figure~\ref{fig:clustering result}, and some generated part-focused images for $N_{part} = 3, 5$ are shown in Figure~\ref{fig:generated part-focused images}.  It can be seen that our part detection method works well for various number of parts.

\subsection{Choice of Parameters}
\label{subsec: parameter choice}
The parameters we need to specify for our proposed system include the set of layers used to detect the object ($\mathcal{S}^{layer\_idx}_{object}$) and the parts ($\mathcal{S}^{layer\_idx}_{parts}$), the number of parts we want to detect ($N_{part}$), and the thresholds used to binarize the masks ($T_{object}$ and $T_{parts}$). Our method is robust to varying $T_{object}$ and  $T_{parts}$ as we will see in Sec.~\ref{sec:Experiments}. In the next two subsections, we will provide some insights on choosing $\mathcal{S}^{layer\_idx}_{object}$, $\mathcal{S}^{layer\_idx}_{parts}$ and $N_{part}$. 

\subsubsection{Choice of $\mathcal{S}^{layer\_idx}_{object}$ and $\mathcal{S}^{layer\_idx}_{parts}$}
\label{subsubsec: set choice}
\begin{figure}[h!]
 \centering
 \includegraphics[width=0.1\textwidth, height=0.1\textwidth]{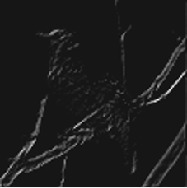}
 \includegraphics[width=0.1\textwidth, height=0.1\textwidth]{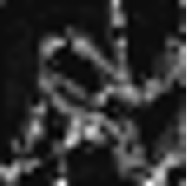}
 \includegraphics[width=0.1\textwidth, height=0.1\textwidth]{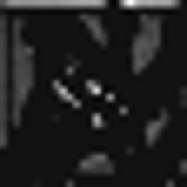} 
 \includegraphics[width=0.1\textwidth, height=0.1\textwidth]{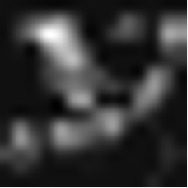}
 \includegraphics[width=0.1\textwidth, height=0.1\textwidth]{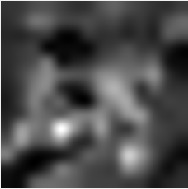}
 \includegraphics[width=0.1\textwidth, height=0.1\textwidth]{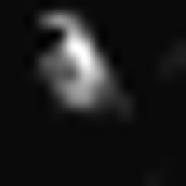} 
 \includegraphics[width=0.1\textwidth, height=0.1\textwidth]{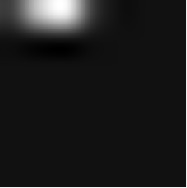} \\
 \includegraphics[width=0.1\textwidth, height=0.1\textwidth]{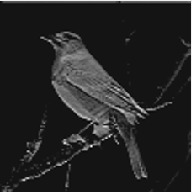}
 \includegraphics[width=0.1\textwidth, height=0.1\textwidth]{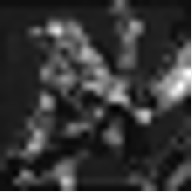}
 \includegraphics[width=0.1\textwidth, height=0.1\textwidth]{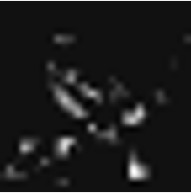} 
 \includegraphics[width=0.1\textwidth, height=0.1\textwidth]{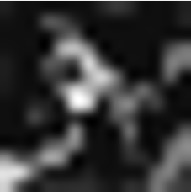}
 \includegraphics[width=0.1\textwidth, height=0.1\textwidth]{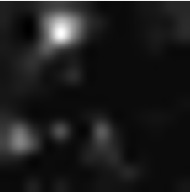}
 \includegraphics[width=0.1\textwidth, height=0.1\textwidth]{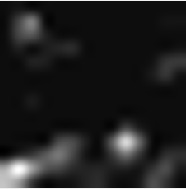} 
 \includegraphics[width=0.1\textwidth, height=0.1\textwidth]{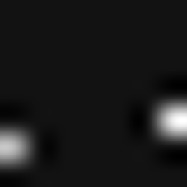} \\
 \includegraphics[width=0.1\textwidth, height=0.1\textwidth]{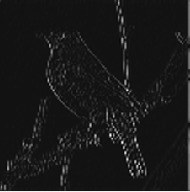}
 \includegraphics[width=0.1\textwidth, height=0.1\textwidth]{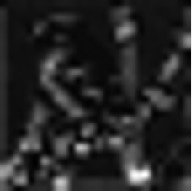}
 \includegraphics[width=0.1\textwidth, height=0.1\textwidth]{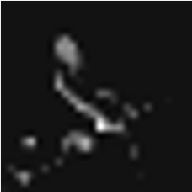} 
 \includegraphics[width=0.1\textwidth, height=0.1\textwidth]{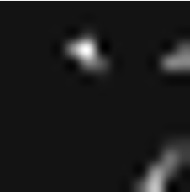}
 \includegraphics[width=0.1\textwidth, height=0.1\textwidth]{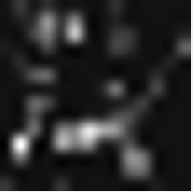}
 \includegraphics[width=0.1\textwidth, height=0.1\textwidth]{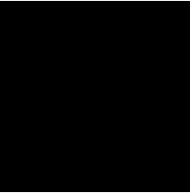} 
 \includegraphics[width=0.1\textwidth, height=0.1\textwidth]{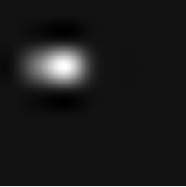} \\
 \includegraphics[width=0.1\textwidth, height=0.1\textwidth]{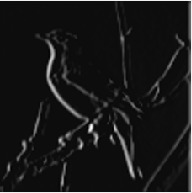}
 \includegraphics[width=0.1\textwidth, height=0.1\textwidth]{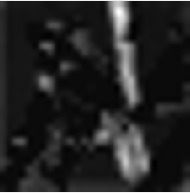}
 \includegraphics[width=0.1\textwidth, height=0.1\textwidth]{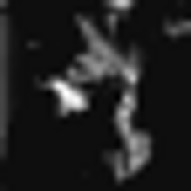} 
 \includegraphics[width=0.1\textwidth, height=0.1\textwidth]{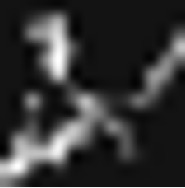}
 \includegraphics[width=0.1\textwidth, height=0.1\textwidth]{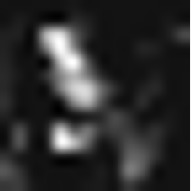}
 \includegraphics[width=0.1\textwidth, height=0.1\textwidth]{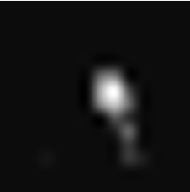} 
 \includegraphics[width=0.1\textwidth, height=0.1\textwidth]{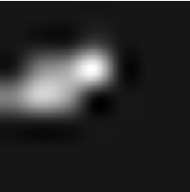} \\
 \includegraphics[width=0.1\textwidth, height=0.1\textwidth]{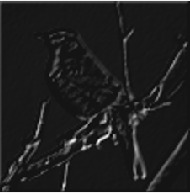}
 \includegraphics[width=0.1\textwidth, height=0.1\textwidth]{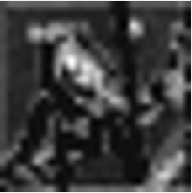}
 \includegraphics[width=0.1\textwidth, height=0.1\textwidth]{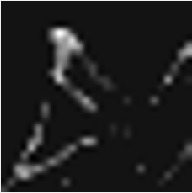} 
 \includegraphics[width=0.1\textwidth, height=0.1\textwidth]{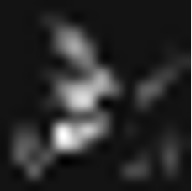}
 \includegraphics[width=0.1\textwidth, height=0.1\textwidth]{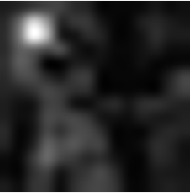}
 \includegraphics[width=0.1\textwidth, height=0.1\textwidth]{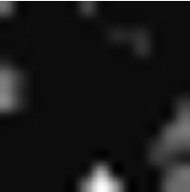} 
 \includegraphics[width=0.1\textwidth, height=0.1\textwidth]{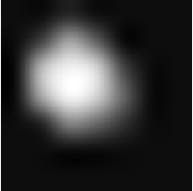} \\
 \includegraphics[width=0.1\textwidth, height=0.1\textwidth]{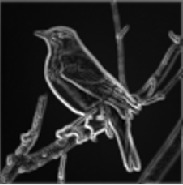}
 \includegraphics[width=0.1\textwidth, height=0.1\textwidth]{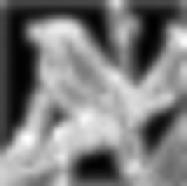}
 \includegraphics[width=0.1\textwidth, height=0.1\textwidth]{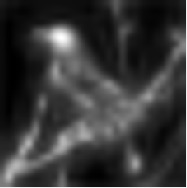} 
 \includegraphics[width=0.1\textwidth, height=0.1\textwidth]{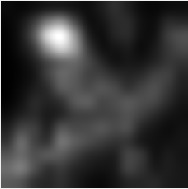}
 \includegraphics[width=0.1\textwidth, height=0.1\textwidth]{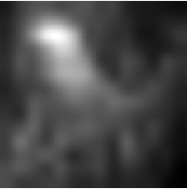}
 \includegraphics[width=0.1\textwidth, height=0.1\textwidth]{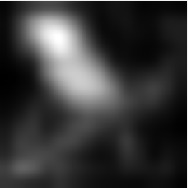} 
 \includegraphics[width=0.1\textwidth, height=0.1\textwidth]{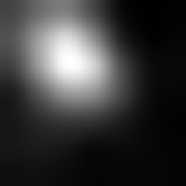} \\
 \caption{Some hidden layer feature maps in GoogLeNet. Detailed description can be found in the text.}
 \label{fig:sample maps}
\end{figure}

Part detectors that only check the local texture without other constraints may lead to false alarm in the background.  For example, when seeing a single black dot, even humans may not be able to tell an eye from a stone.  Seeing the whole object first provides the context to focus on specific parts.  Consequently, our algorithm first detects the object region and then uses it to constrain the subsequent part detection task.  For both object and part detection, one result is given by one mask which is the normalized version of the sum of some selected hidden layer feature maps. $\mathcal{S}^{layer\_idx}_{object}$ and $\mathcal{S}^{layer\_idx}_{parts}$ specify the layers whose feature maps are most likely to contribute to object detection and part detection, respectively. Figure~\ref{fig:sample maps} shows some resized feature maps from different hidden layers in the GoogLeNet \cite{GoogLeNet}, which has $22$ layers with parameters or $27$ layers if the pooling layers are also counted. (The original image is the first example in Figure~\ref{fig:sample masks}.) Each column in this figure contains five feature maps of one layer and the last one is the normalized sum of all the maps in that layer. Considering the GoogLeNet with $22$ layers, from left to right shows seven of them which are the $\nth{1}$, $\nth{8}$, $\nth{9}$, $\nth{11}$, $\nth{13}$, $\nth{15}$, and $\nth{18}$ layers. 

From the sample feature maps shown above, we can see that the lower layers tend to always respond to local patterns and the activated regions are scattered.  An extreme case is the $\nth{1}$ layer maps which respond strongly to the edges of the bird and the branches alike and hence the sum map of this layer is like an edge detection result.  As we move to higher levels, the activated regions become more compact in each map and the sum map is more focused on the object as a whole.  This also justifies our part detection procedure described in Subsec.~\ref{subsec:Part detection}, by using lower layer feature maps selected by object detection result form higher layer, we leverage both the precision of lower layer and the robustness of higher layer.

Another observation is that in every layer, we can often find some feature maps that are activated by the background, and the object and parts are often partially detected in one map. However, the simple summation and normalization operations can sufficiently suppress the spurious activations and integrate the partial detections into a complete one.  The boundary is complete in the $\nth{1}$ layer normalized sum map.  As we move to higher layers, the background branches fade away in their normalized sum maps.  The whole object region is clearly detected in the sum map of the $\nth{18}$ layer although its individual maps have both false alarm and incomplete object detection.

Since $\mathcal{S}^{layer\_idx}_{object}$ specifies the layer(s) whose features are most likely to be activated by object-level patterns, it is reasonable to pick the layer(s) high enough to perceive global view of the input image. From Figure.~\ref{fig:sample maps} we can see that the sum maps in lower layers include more background activation and consequently the object detection is not tight. Similar consideration applies to the choice of $\mathcal{S}^{layer\_idx}_{parts}$, where we hope to see that the activations are compact and fine enough to highlight one object part or part of it.  Considering how we select hidden layer feature maps to detect parts in Subsec.~\ref{subsec:Part detection} together with the results in Figure.~\ref{fig:sample maps}, we can see that too high layers will not distinguish the parts but the object, while the activations in low layers' feature maps tend to contain multiple disconnected regions instead of one, so few of them are valid to detect parts, making the detection result not robust.  The choice of $\mathcal{S}^{layer\_idx}_{object}$ and $\mathcal{S}^{layer\_idx}_{parts}$ for the CNNs we use are listed in Table~\ref{tab:layer choice}.  We have tried similar choices and they also works.

\begin{table}[h]
\begin{center}
{\small
\begin{tabular}{|l||p{2.6cm} | p{2.6cm}|}
\hline
\multicolumn{1}{|c||}{CNN}            & $\mathcal{S}^{layer\_idx}_{object}$       & $\mathcal{S}^{layer\_idx}_{parts}$ \\
\hline\hline
GoogLeNet \cite{GoogLeNet}           &  inception\_4e/output, inception\_5a/output & inception\_4d/output, inception\_4e/output \\
\hline
VGG19 \cite{VGG}                     &  conv5\_4                                  & conv5\_2, conv5\_3      \\
\hline
VGG-CNN-S \cite{chatfield2014return}  &  conv5                                    & conv4, conv5         \\
\hline
\end{tabular}
}
\end{center}
\caption{The choice of $\mathcal{S}^{layer\_idx}_{object}$ and $\mathcal{S}^{layer\_idx}_{parts}$ for GoogLeNet, VGG19 and VGG-CNN-S.}
\label{tab:layer choice}
\end{table}

\subsubsection{Choice of $N_{part}$}
\label{subsubsec: set N part}
As described in Algorithm \ref{alg:mask generation}, we cluster the selected feature maps' centroids by $k$-means and we choose $k$ according to Davies-Bouldin criterion \cite{davies1979cluster} and Silhouette criterion \cite{kaufmanpj,rousseeuw1987silhouettes}.

The Davies-Bouldin criterion \cite{davies1979cluster} is based on the ratio of the distances between samples of the same and different clusters.  The Davies-Bouldin index ($DB$) of a clustering with $k$ groups is defined as:

\begin{equation}
 DB = \frac{1}{k} \sum_{i=1}^k max_{j \neq i} \{D_{i,j}\}
\end{equation}

\begin{equation}
 D_{i,j} = \frac{\textit{within-cluster distance}}{\textit{between-cluster distance}} = \frac{\bar{d}_i + \bar{d}_j}{d_{i,j}}
\end{equation}

$\bar{d}_i$ is the average distance between points and their own centroid in the i\textsuperscript{th} cluster.  $d_{i,j}$ is the distance between the centroids of the i\textsuperscript{th} and j\textsuperscript{th} clusters.

The maximum value of $D_{i,j}$ over $j$ gives a score of the most ambiguous case for group $i$ and $DB$ is obtained by averaging the worst case scores of all clusters.  The smaller the $DB$ is, the better the clustering result. 

\begin{figure}[h!!]
 \centering
 \includegraphics[width=\textwidth]{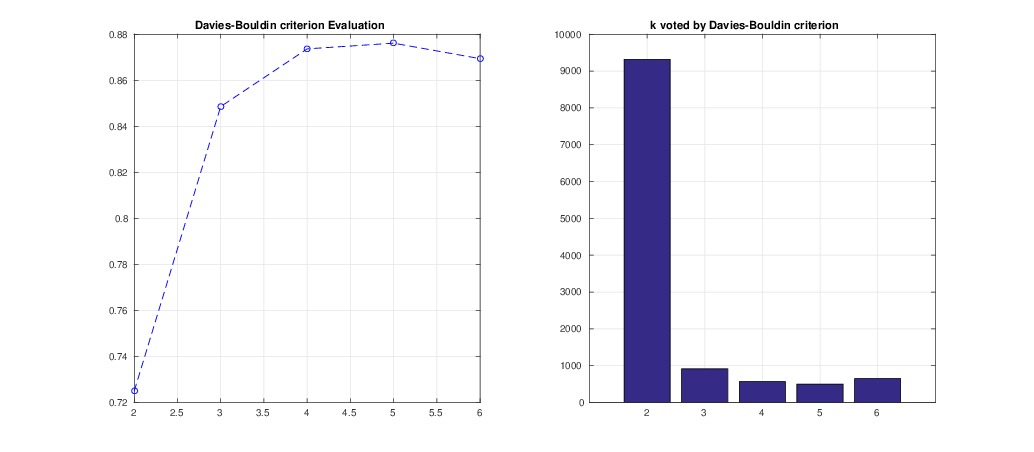}
 \caption{Different choice of $k$ in k-means clustering measured by Davies-Bouldin criterion on Caltech-UCSD Birds-200-2011 dataset}
 \label{fig: DB CUB}
\end{figure}

\begin{figure}[h!!]
 \centering
 \includegraphics[width=\textwidth]{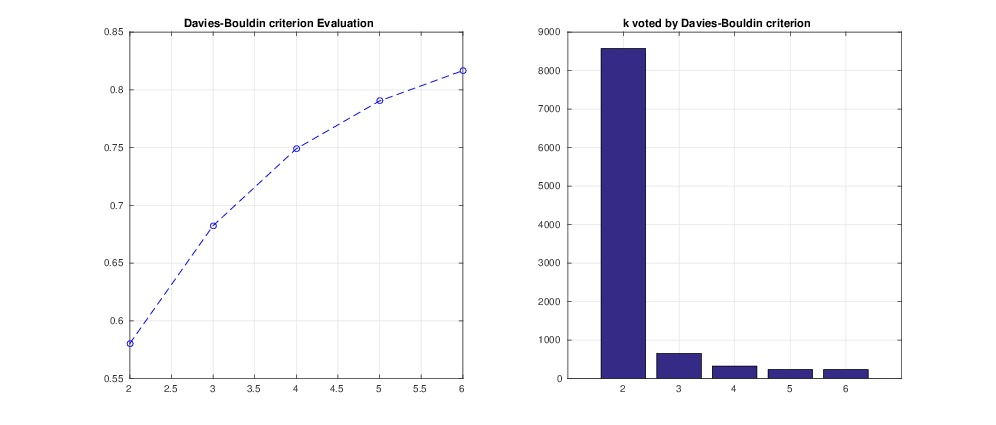}
 \caption{Different choice of $k$ in k-means clustering measured by Davies-Bouldin criterion on FGVC-Aircraft dataset}
 \label{fig: DB FGVC}
\end{figure}

The examples of using Davies-Bouldin criterion to measure our k-means clustering with different $k$ ($N_{part}$) on Caltech-UCSD Birds-200-2011~\cite{WelinderEtal2010} and FGVC-Aircraft~\cite{fgvcaircraft} datasets are shown in Figure~\ref{fig: DB CUB} and Figure~\ref{fig: DB FGVC}.  The left figure shows the mean Davies-Bouldin index of all the images in the dataset and the right figure shows the histogram of best the $k$ choice.  The cars~\cite{krause20133d} and Stanford dogs~\cite{khosla2011novel} datasets give similar results where $k = 2$ is preferred.

The silhouette value $Si$ for the i\textsuperscript{th} data point is defined as:

\begin{equation}
 Si = \frac{bi-ai}{max(ai,bi)}
\end{equation}

where $ai$ is the average distance from the i\textsuperscript{th} point to the other points within the same cluster and $bi$ is the minimum average distance from the i\textsuperscript{th} point to the points in different clusters, and the minimization is taken over all the other clusters.

It can be seen that $-1 \leq Si \leq 1$  and the higher the silhouette value the safer the i\textsuperscript{th} point is in its own cluster.  If most points have a high silhouette value, then the clustering solution is appropriate.  So a high average value of $Si$ over all the data is an indication of good clustering.

\begin{figure}[h!!]
 \centering
 \includegraphics[width=\textwidth]{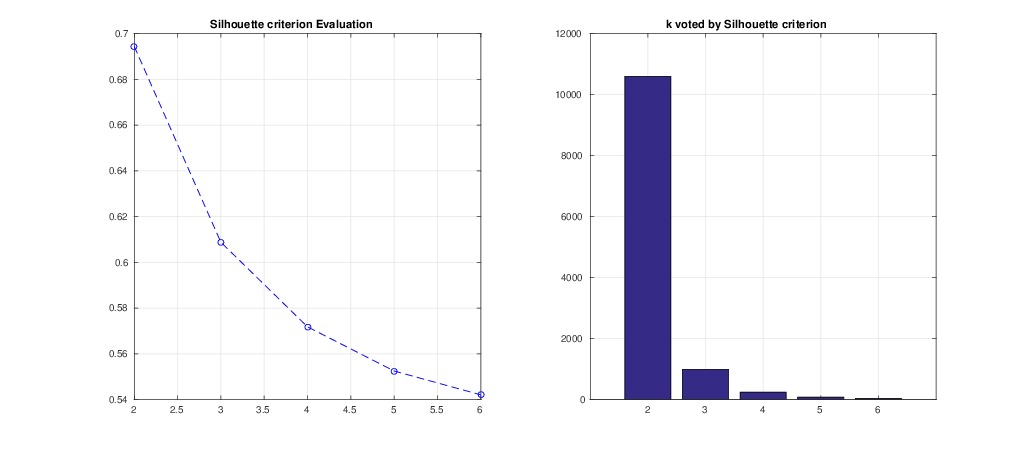}
 \caption{Different choice of $k$ in k-means clustering measured by Silhouette criterion on Caltech-UCSD Birds-200-2011 dataset}
 \label{fig: Si CUB}
\end{figure}

\begin{figure}[h!!]
 \centering
 \includegraphics[width=\textwidth]{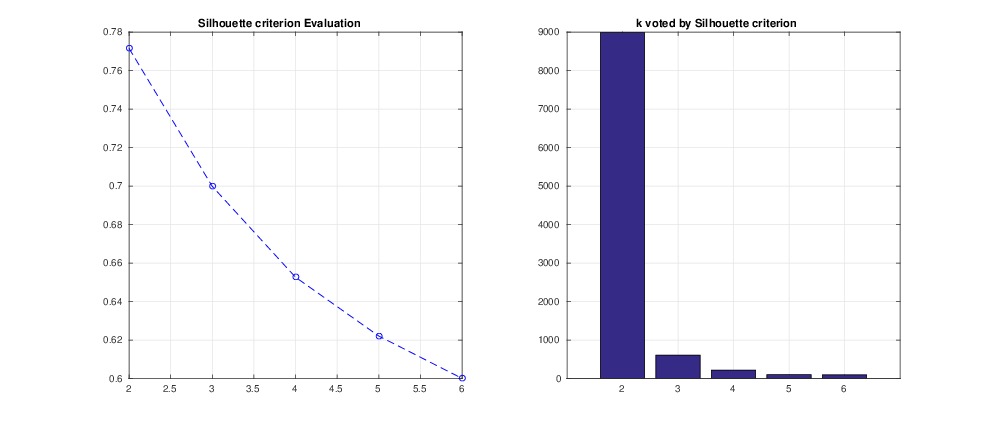}
 \caption{Different choice of $k$ in k-means clustering measured by Silhouette criterion on FGVC-Aircraft dataset}
 \label{fig: Si FGVC}
\end{figure}

The results of applying Silhouette criterion on Caltech-UCSD Birds-200-2011~\cite{WelinderEtal2010} and FGVC-Aircraft~\cite{fgvcaircraft} datasets are shown in Figure~\ref{fig: Si CUB} and Figure~\ref{fig: Si FGVC}.  The left figure shows the mean Silhouette value of all the images in the dataset and the right figure shows the histogram of best $k$ choice.  The cars~\cite{krause20133d} and Stanford dogs~\cite{khosla2011novel} datasets again give similar results.

\begin{figure}[h!!]
 \centering
 \includegraphics[width=\textwidth]{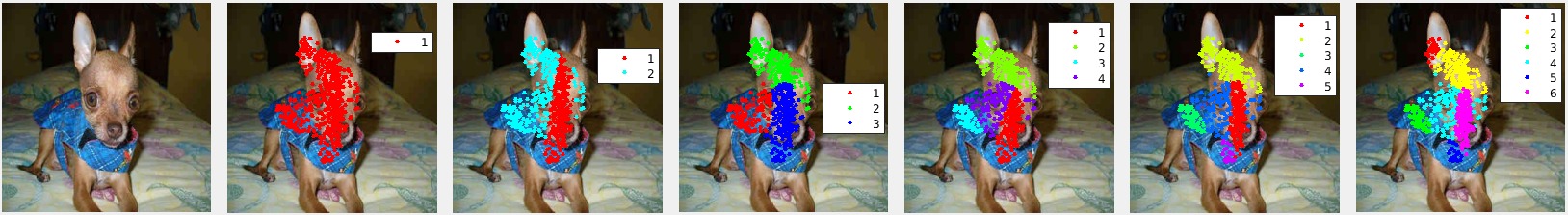}
 \includegraphics[width=\textwidth]{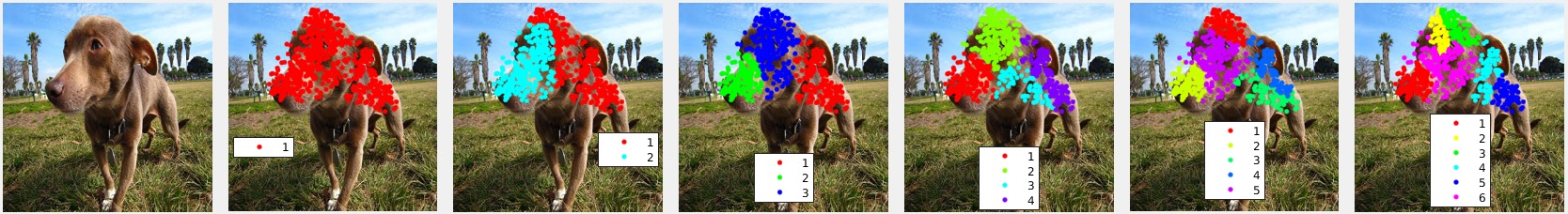}
 \caption{Clustering samples with $N_{part} = 1, 2, 3, 4, 5, 6$ of Stanford dogs \cite{khosla2011novel} datasets.}
 \label{fig: dog part}
\end{figure}

It can be seen that both criteria suggest $k=2$ is preferred by the majority.  However, the above mentioned two criteria are not appliable to $k=1$ case.  We found during the experimets that for cars~\cite{krause20133d} and Stanford dogs~\cite{khosla2011novel} datasets, $k=1$ is more suitable.  This is due to the property of the datasets.  Two typical images from Stanford dogs~\cite{khosla2011novel} dataset and the clustering of the selected feature maps' centroids are shown in Figure~\ref{fig: dog part} where most of the feature maps are focused on the face of the dogs for the following reasons: First unlike birds, dogs do not have colorful feather. The furs of dogs are mostly white, brown and black without special texture.  Second, most of the dogs in the images are pets that are facing the camera, with their bodies occluded by the head, human hands or clothes.  Thus, dogs are mainly distinguished by their heads, so only one part is focused.  Similarly, cars~\cite{krause20133d} of different models differ from each other mainly by their shapes of a large chunk rather than special local texture.    

The choice of $N_{part}$ is further justified by the classification results in Subsec.~\ref{subsec:explain away ensemble}.  A small $N_{part}$ with each part-focused image containing a reasonable large portion of the object is actually quite preferable considering some practical issues of our fine-grained categorization framework.  The purpose of part detection is to facilitate fine-grained local feature extraction by inputing cropped part-focused images into CNNs.  As discussed previously, patterns are better interpreted in context. Even if the eye of a bird is precisely detected, cropping just a black dot does not provide very useful features, but including the feather around the eye can often help a lot.   Another reason for preferring a small $N_{part}$ is that in many cases it is hard to identify the exact boundary between parts, e.g., between the head and neck, neck and chest, belly and back, etc.  Moreover, sometimes one part may partially occlude another part, e.g., the wings covering the body.  Besides, the order in which we concatenate the features of the parts is based on the size of the part region detected.  A consequence of having a larger $N_{part}$ would be increasing the chance of getting parts with similar size and hence mis-ordering them. A practical consideration is that more parts require more memory in the integrated CNN.  Last but not least, if the problem setting changes and part annotations are used, our method can still play an important role, e.g.,generating high-quality part proposals, etc.

\subsection{Part-Focused Image Generation}
\label{subsubsec: part-pose}
\begin{figure}[h!!]
 \centering
 \includegraphics[width=0.7\textwidth, height=0.078\textwidth]{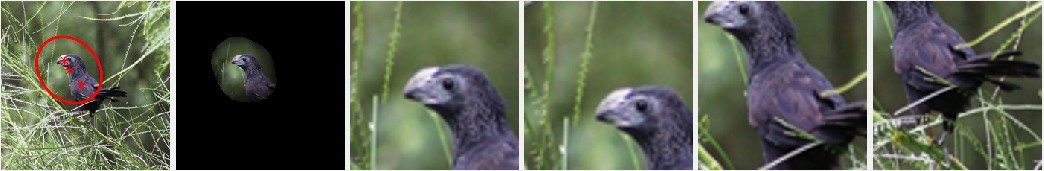}
 \includegraphics[width=0.7\textwidth, height=0.078\textwidth]{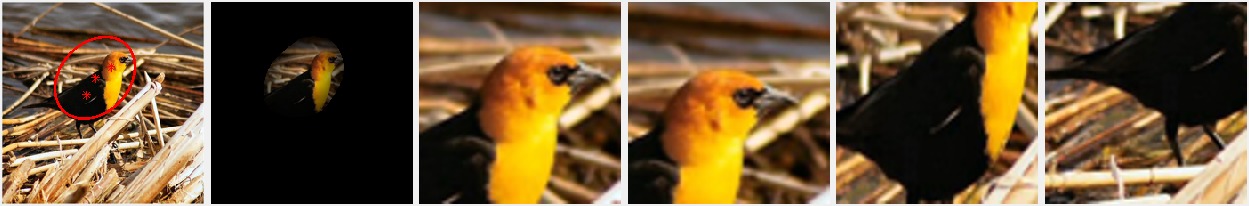}
 \includegraphics[width=0.7\textwidth, height=0.078\textwidth]{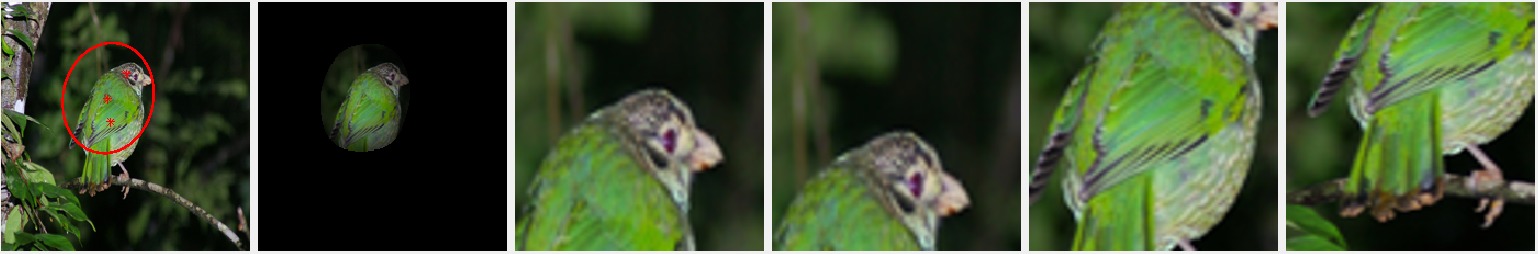}
 \includegraphics[width=0.7\textwidth, height=0.078\textwidth]{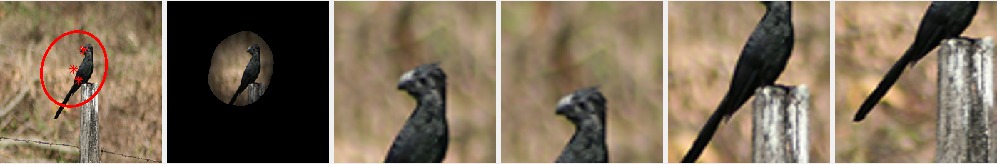}
 \includegraphics[width=0.7\textwidth, height=0.078\textwidth]{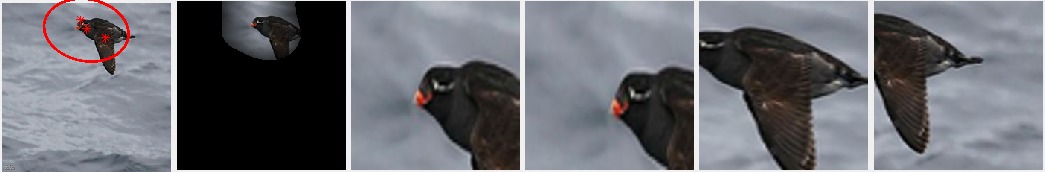}
 \includegraphics[width=0.7\textwidth, height=0.078\textwidth]{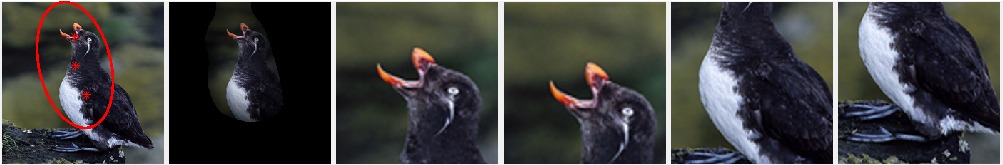}
 \includegraphics[width=0.7\textwidth, height=0.078\textwidth]{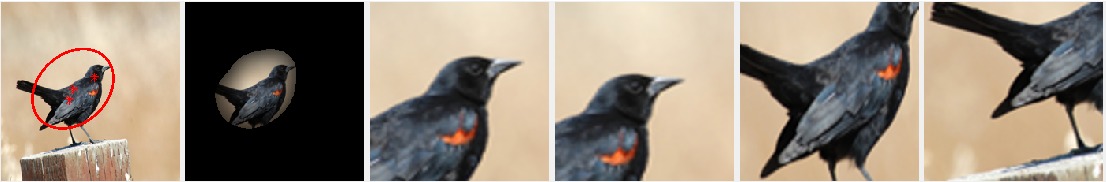}
 \includegraphics[width=0.7\textwidth, height=0.078\textwidth]{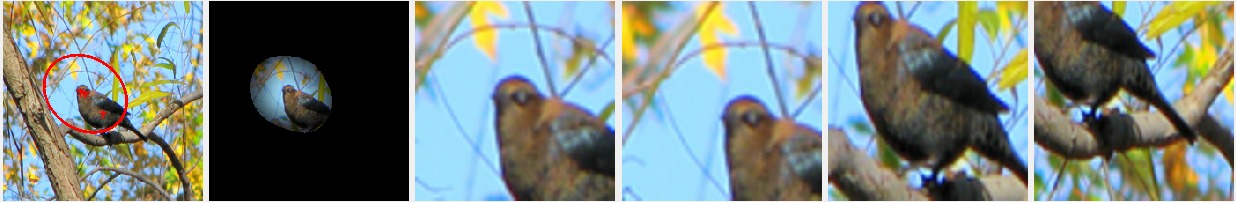}
 \includegraphics[width=0.7\textwidth, height=0.078\textwidth]{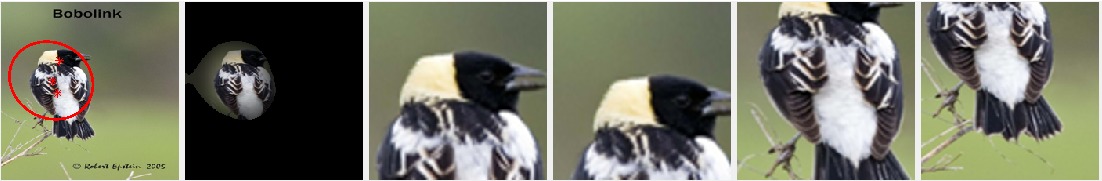}
 \includegraphics[width=0.7\textwidth, height=0.078\textwidth]{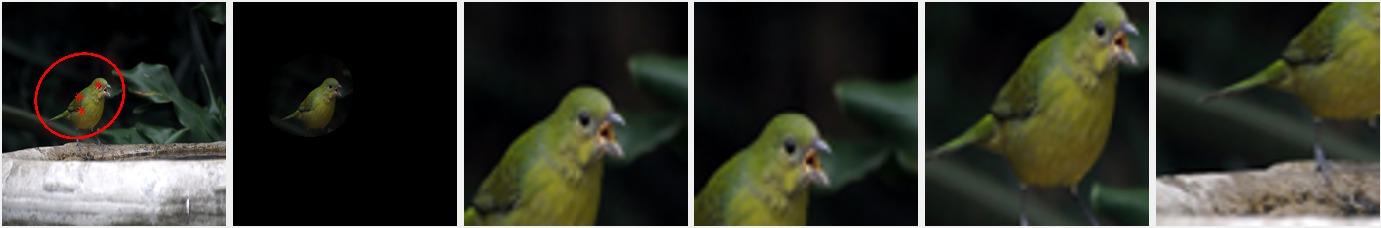}
 \caption{Some examples of part-focused image generation with and without pose adjustment of the cropping. Each row shows one example.  The first image of each row is the original image, with the detected object region represented roughly by an ellipse. The weighted centroids of the object region and two part regions are marked with red stars. The third and fifth images of each row are the two part-focused images generated by directly cropping image patches around the detected part regions. The fourth and sixth images are the part-focused images generated after adjusting the position and size of the cropped image patches along the direction from the centroid of the object to the centroids of the two parts (using the pose estimation represented by two vectors).}
 \label{fig: part-focused image generation}
\end{figure}

Based on the saliency mask for each object part, the most intuitive way to generate a part-focused image is to directly crop a patch around the salient region.  However, we observe that sometimes both the object and part detection results are biased towards the regions which have complex patterns and are crucial for classification while some less discriminative regions with simpler patterns are underestimated. This is because we aim at achieving robust detection by combining many feature maps but the regions with relatively simple and indistinctive patterns have few feature maps voting for them.  Figure~\ref{fig: part-focused image generation} (better viewed in color) demonstrates this phenomenon in applying GoogLeNet~\cite{GoogLeNet} on Caltech-UCSD Birds-200-2011~\cite{WelinderEtal2010} with $N_{part}=2$. The first image of each row is the original image, with the detected object region represented roughly by an ellipse. The weighted centroids of the object region and two part regions are marked with red stars. We can see that the two parts detected correspond roughly to the head and body, but the salient region and the centroid of the object are biased towards the head. The third and fifth images of each row are the two part-focused images generated by directly cropping image patches around the active region in part detection masks. It can be seen that part of the head is sometimes included in the body-focused image while part of the tail is often ignored.

Another inspiring observation from Figure~\ref{fig: part-focused image generation} is that although the detection results are biased, they are sufficient to indicate the pose of the bird. Thus, the bias can be easily compensated by shifting and extending the cropped region along the pose of the bird, e.g, in order to include the mistakenly ignored tail, we can simply shift and extend the cropped image patch to the pointing direction of the tail.

\begin{figure}[h!]
 \centering
 \includegraphics[width=0.11\textwidth]{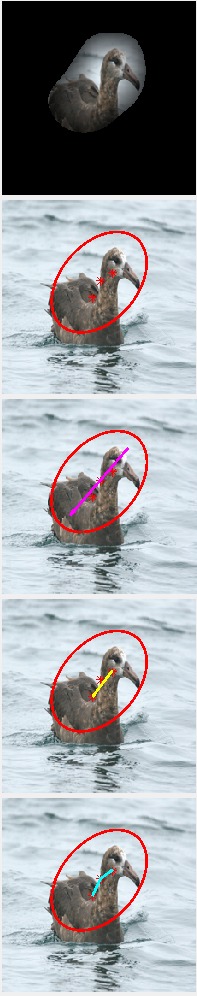} 
 \includegraphics[width=0.11\textwidth]{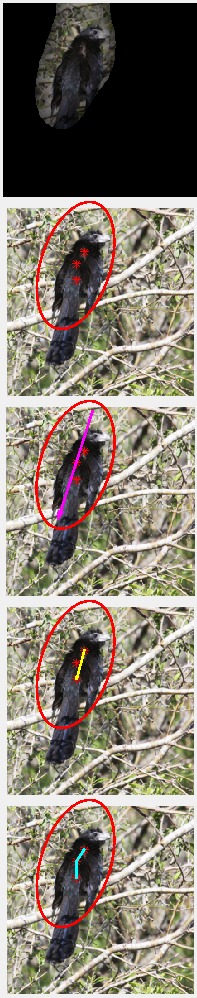} 
 \includegraphics[width=0.11\textwidth]{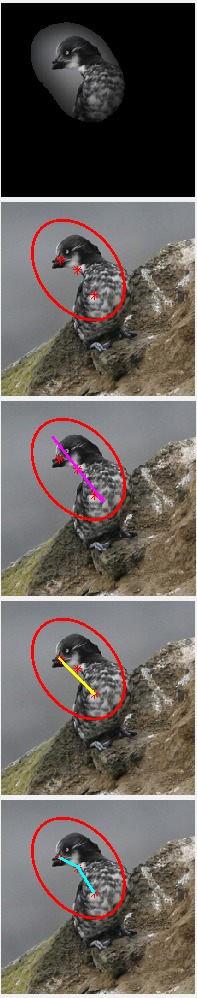} 
 \includegraphics[width=0.11\textwidth]{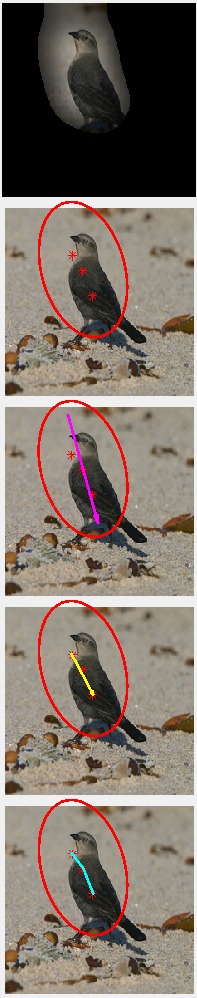}
 \includegraphics[width=0.11\textwidth]{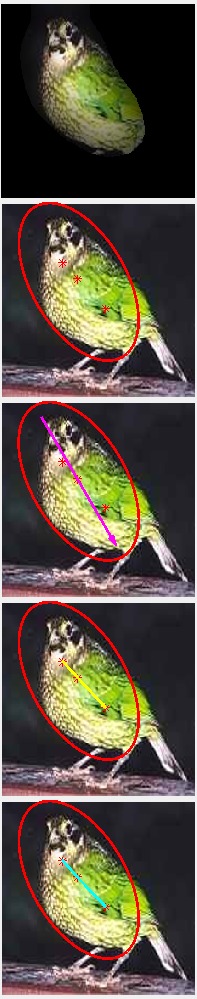}
 \includegraphics[width=0.11\textwidth]{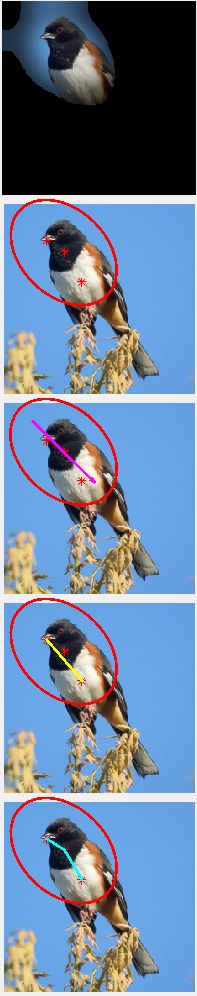}
 \includegraphics[width=0.11\textwidth]{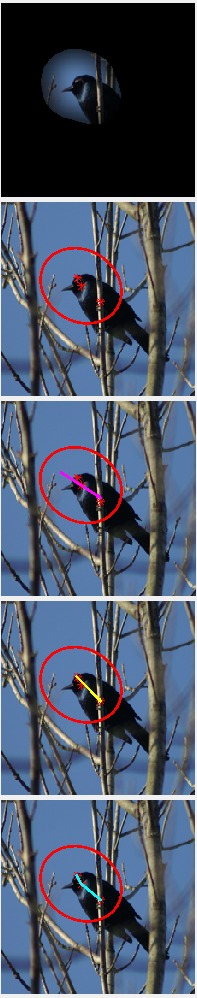}
 \includegraphics[width=0.11\textwidth]{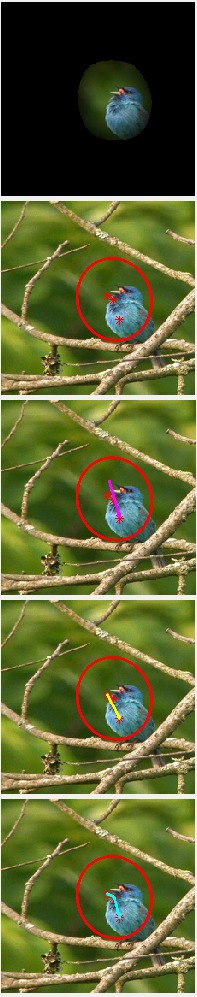}
 \caption{Pose estimation obtained by the three ways described in the text. Each column shows one example. From top to bottom: the first one shows the soft-masked image; the second one is the original image with the detected object region represented roughly by an ellipse and the weighted centroids of the object region and two part regions are marked with red stars; the third one visualizes the pose obtained by the vector between the two focal points of the ellipse; the fourth one visualizes the pose obtained by the vector between the centroid of the head region and that of the body region; the fifth one visualizes the pose obtained by the two vectors from the centroid of the object, with one ending at the centroid of the head while the other at that of the body.}
 \label{fig: pose}
\end{figure}

For the case of $N_{part}=2$, given the ellipse which roughly contains the bird, the centroid of the bird and those of its head and body, we can represent the pose of the bird by one vector pointing from the head to the body (or the opposite direction), or two vectors from the center of the bird to the two parts. Intuitively, we have two choices of the pose represented by a single vector: the vector that starts from the ellipse's focal point closer to the centroid of the head and ends at the other focal point; the vector that starts from the centroid of the head and ends at the centroid of the body.  There is one choice of the pose represented by two vectors: both vectors start at the centroid of the object, with one ending at the centroid of the head while the other ending at the centroid of the body. The three pose choices are illustrated in Figure~\ref{fig: pose} (better viewed in color) with each column showing one example.  We can see that the three methods give very consistent pose estimation results. We have tried all these three versions of poses to adjust part-focused image  generation and they indeed offer similar results. In our experiment we use the two-vector version, which can be extended to more parts cases intuitively.

\begin{figure}[h!]
 \centering
 \includegraphics[width=0.2\textwidth, height=0.2\textwidth]{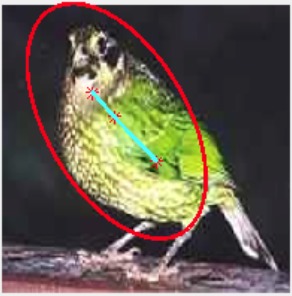}
 \includegraphics[width=0.02\textwidth, height=0.15\textwidth]{white_space.jpg}
 \includegraphics[width=0.25\textwidth, height=0.15\textwidth]{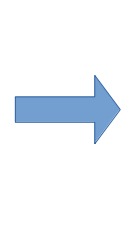}
 \includegraphics[width=0.02\textwidth, height=0.15\textwidth]{white_space.jpg}
 \includegraphics[width=0.45\textwidth, height=0.3\textwidth]{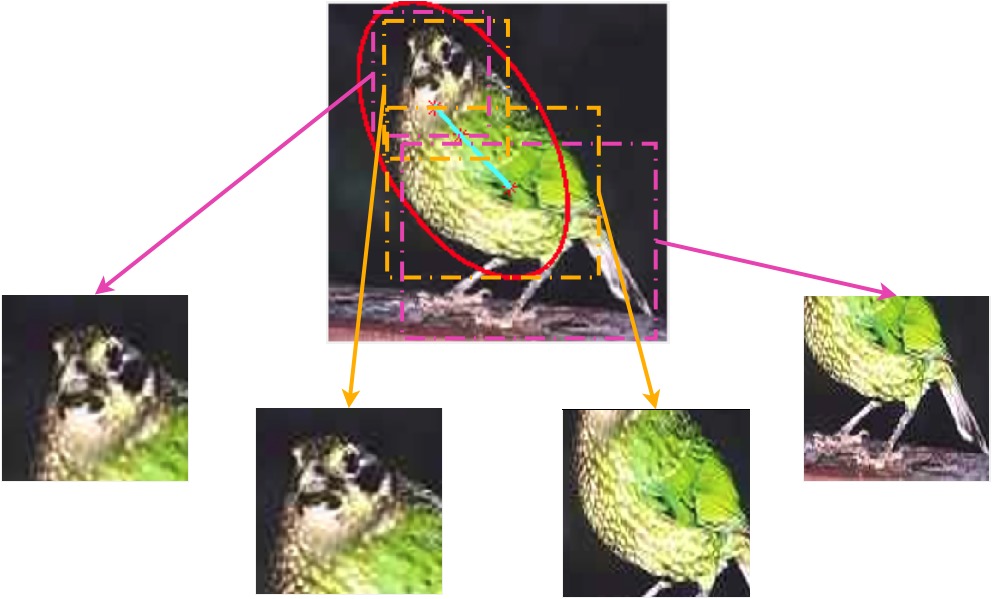}
 \caption{Illustration of shifted cropping based on the pose estimation result of $N_{part} = 2$. The orange rectangles are the regions of directly cropped patches according to the part detection mask, and the pink ones are the shifted cropped regions according to the estimated pose.}
 \label{fig:pose adjust}
\end{figure}

In our system, we shift the cropped image patches along the pose to compensate potential bias and augment the training images. Figure~\ref{fig:pose adjust} illustrates this procedure and the cropped image patches after adjustment are the forth and sixth columns in Figure~\ref{fig: part-focused image generation}.

\subsection{Integration of Specialized CNNs}
After the object-focused and part-focused images are generated, we construct one more CNN initialized with that previously trained using the raw images for each part. We fine-tune the CNNs for the object parts using the corresponding part-focused images so the new CNNs are specialized to extract discriminative features from different parts.  To integrate different CNNs specialized for object-level and part-level features, we concatenate the feature vectors followed by a dropout layer with dropout ratio 0.3, then a fully-connected layer with the out put number matching the number of classes is added.  The integrated CNN is fine-tuned for the last time.  In Figure~\ref{fig:sys}, the newly added module (blue rectangle) comprises the concatenation layer and dropout layer mentioned above. The classifiers in Figure~\ref{fig:sys} are simply a fully-connected layer with the number of output units the same as that of the classes followed by a softmax layer.

\begin{figure}[h!!]
 \centering
 \includegraphics[width=0.85\textwidth]{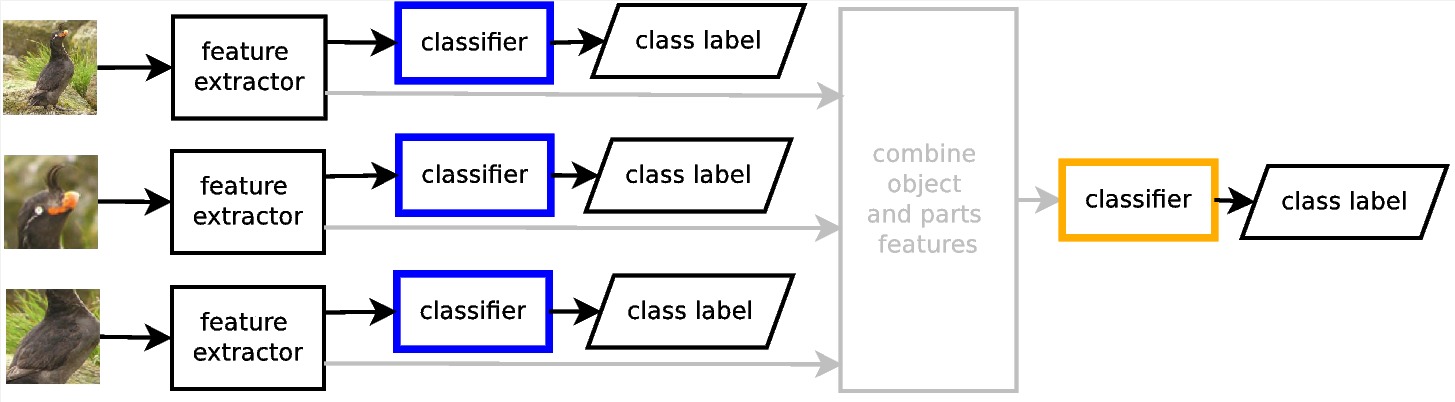}
 \caption{Separate fine-tuning of the three sub-CNNs}
 \label{fig: fine-tune separate}
\end{figure}

\begin{figure}[h!!]
 \centering
 \includegraphics[width=0.85\textwidth]{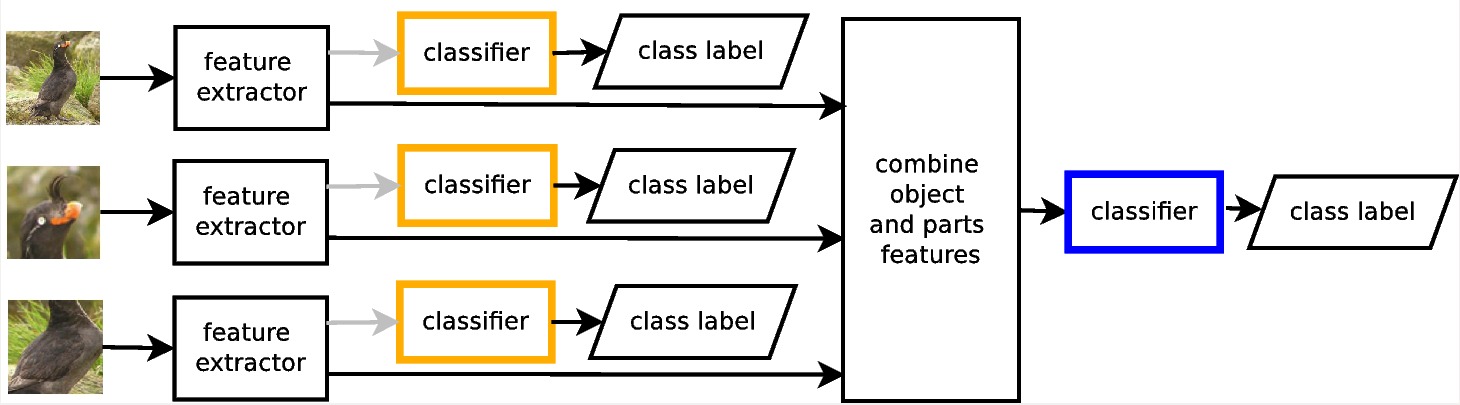}
 \caption{Overall fine-tuning of the integrated CNN}
 \label{fig: fine-tune overall}
\end{figure}

Thus, we specialize our feature extractors but changing the network structure is unappealing.  In fact we only need to construct one CNN for these two phases, and adjust the $loss\_weight$ parameter in \cite{caffe} once.  We design our CNN for training as shown in Figure~\ref{fig: fine-tune separate} and \ref{fig: fine-tune overall} (the same network with different parameter settings shown in different colors).  Compared with Figure~\ref{fig:sys}, the difference is that three more classifiers are plugged in directly after each feature extractor. The strength of supervision imposed by each classifier is adjusted by the $loss\_weight$ parameter in \cite{caffe}. If the $loss\_weight$ of one classifier is set to $0$, the error calculated by this classifier is discarded and no gradient information is propagated to the path below. In Figure~\ref{fig: fine-tune separate} and \ref{fig: fine-tune overall}, the classifiers with $loss\_weight = 0$ are marked by an orange frame, the classifiers with $loss\_weight = 1$ are marked by a blue frame, and the `silent' backward propagation paths are shown in gray. Figure~\ref{fig: fine-tune separate} shows the integrated CNN conducting completely separate fine-tuning for its three specialized feature extractors and Figure~\ref{fig: fine-tune overall} shows the integrated CNN conducting overall fine-tuning based on the errors from only one final classifier. Another option is to set a non-zero $loss\_weight$ for all the classifiers simultaneously.  Then both the overall loss and separate supervision are considered during the update of the feature extractors. In our experiments, we first specialize the three feature extractors as in Figure~\ref{fig: fine-tune separate}, and then fine-tune the overall network with $loss\_weight = 1$ for the final classifier. During the overall fine-tuning, we try both $loss\_weight = 0$ and $loss\_weight = 0.3$ for the classifiers directly following the three feature extractors, and these two strategies achieve almost the same test accuracy.

During the concatenation of the feature vectors of the two parts, we order them according to the sizes of their detected regions.  Although this appears to be too simplistic, our experimental results to be presented in Sec.~\ref{sec:Experiments} show that it works surprisingly well.  When there are only a small number of parts, the different parts usually differ in size, e.g., the head and body of humans or animals.  This scheme may also be generalized to more parts by detecting them in a hierarchical way.  To do so, we first roughly decompose an object into a few major parts.  This is then followed by detecting finer parts within each major part.  Not only does this simplify the identification of detected parts, but it also increases the robustness of the detection results since the finer parts are constrained in their location by the major part to which they belong.  In Subsec~\ref{subsubsec: part-pose}, we mention that the object pose can be estimated based on our object and part detection results, these results can assist in identifying and ordering the parts as well.

\section{Experiments}
\label{sec:Experiments}
In this section, we report the experimental validation of our proposed method.  Subsec~\ref{subsubsec: obj part detection results} shows the object detection and part detection results.  Subsec~\ref{subsubsec: robustness of detection results} uses some examples to illustrate that the detection results are not sensitive to the thresholds $T_{object}$ and $T_{parts}$.  Since our proposed method integrates multiple CNNs, the experiments in Subsec~\ref{subsec:explain away ensemble} are conducted to explain away that the performance gain does not comes from ensemble.  Finally, comparison with our baselines and other methods are listed in Subsec~\ref{subsec:comparison with other methods}.

\begin{figure}[h!]
 \centering
 \includegraphics[width=0.11\textwidth, height=0.1\textwidth]{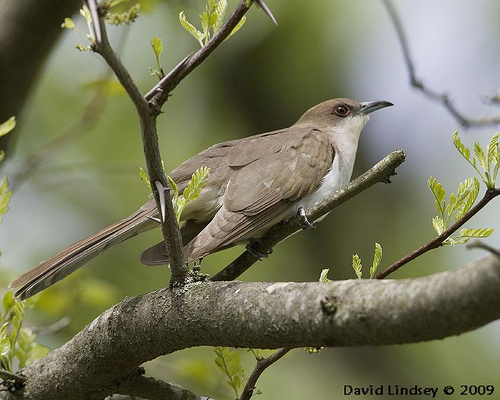}
 \includegraphics[width=0.11\textwidth, height=0.1\textwidth]{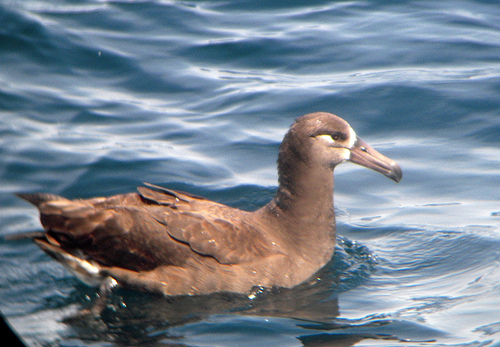}
 \includegraphics[width=0.11\textwidth, height=0.1\textwidth]{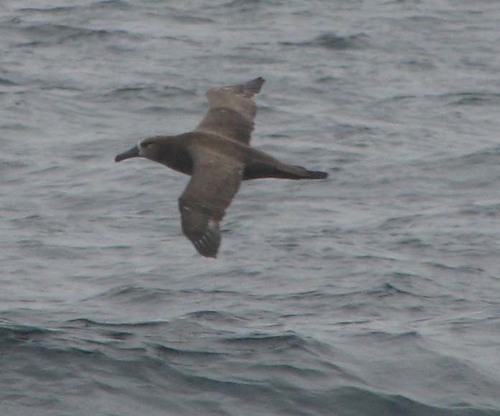}
 \includegraphics[width=0.11\textwidth, height=0.1\textwidth]{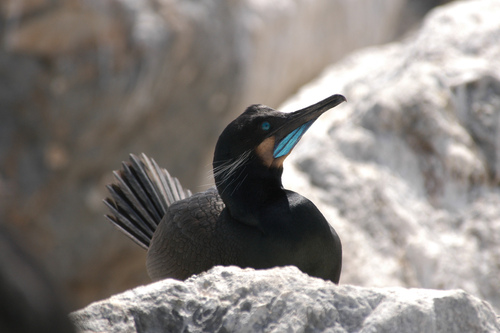}
 \includegraphics[width=0.11\textwidth, height=0.1\textwidth]{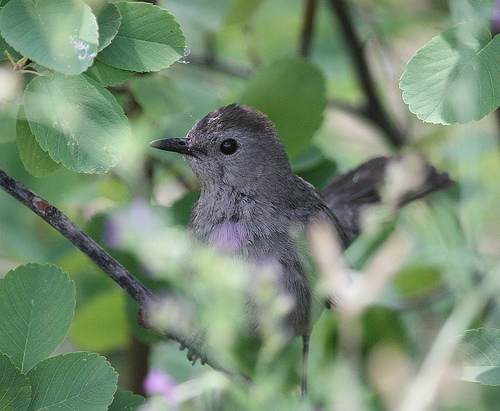}
 \includegraphics[width=0.11\textwidth, height=0.1\textwidth]{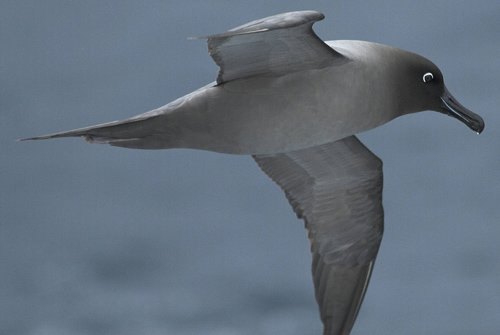}
 \includegraphics[width=0.11\textwidth, height=0.1\textwidth]{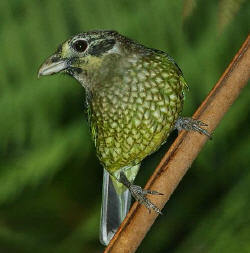} \\
 \includegraphics[width=0.11\textwidth, height=0.1\textwidth]{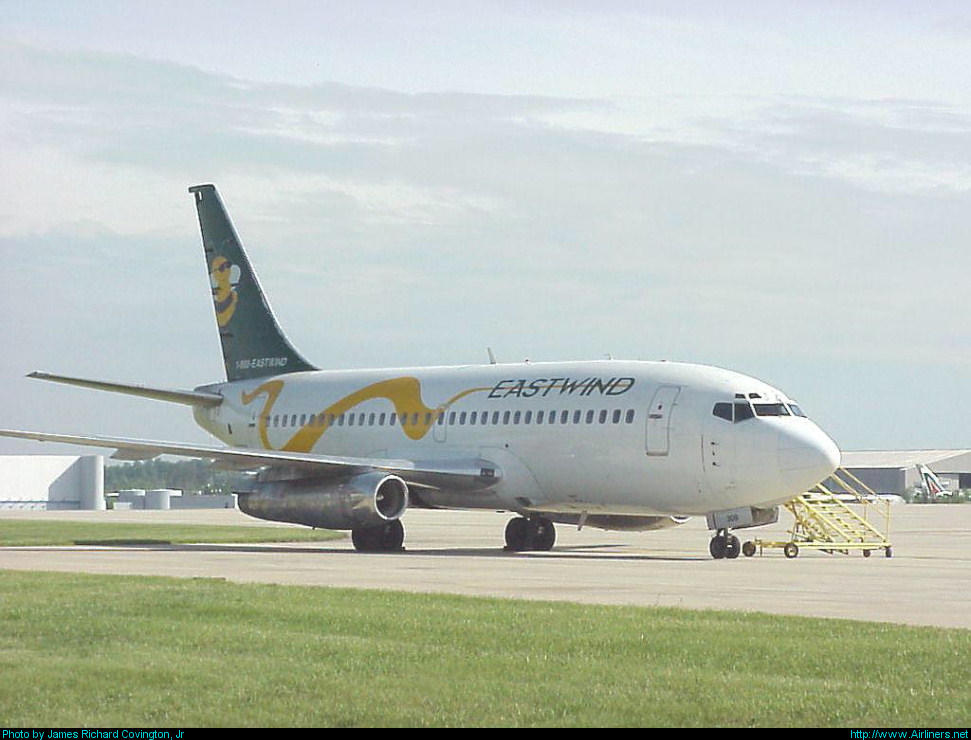}
 \includegraphics[width=0.11\textwidth, height=0.1\textwidth]{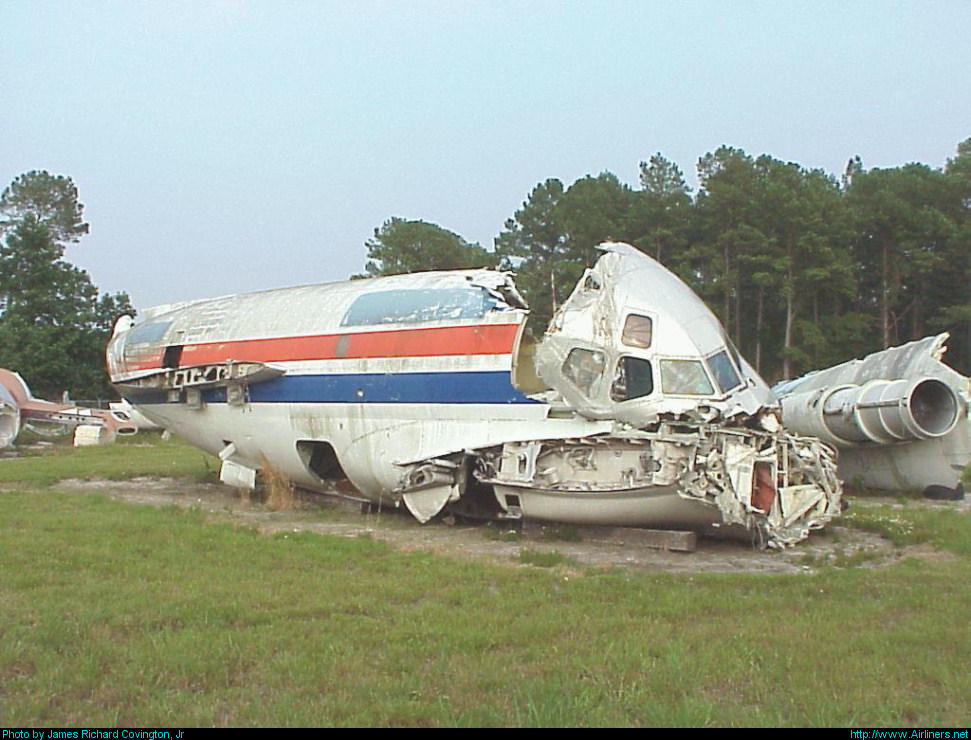}
 \includegraphics[width=0.11\textwidth, height=0.1\textwidth]{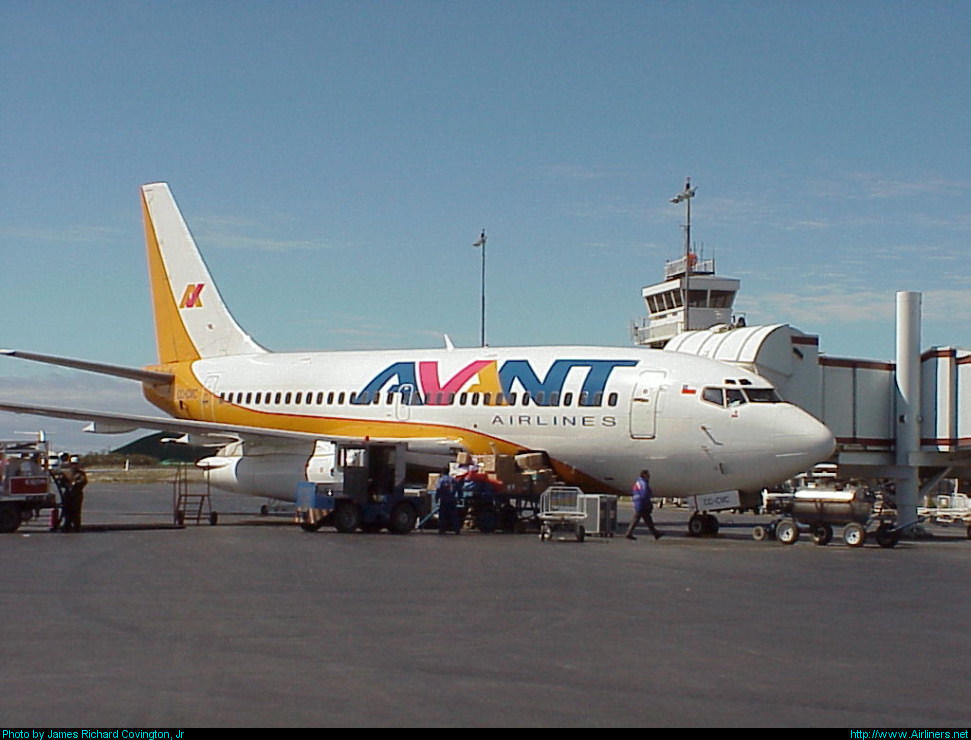}
 \includegraphics[width=0.11\textwidth, height=0.1\textwidth]{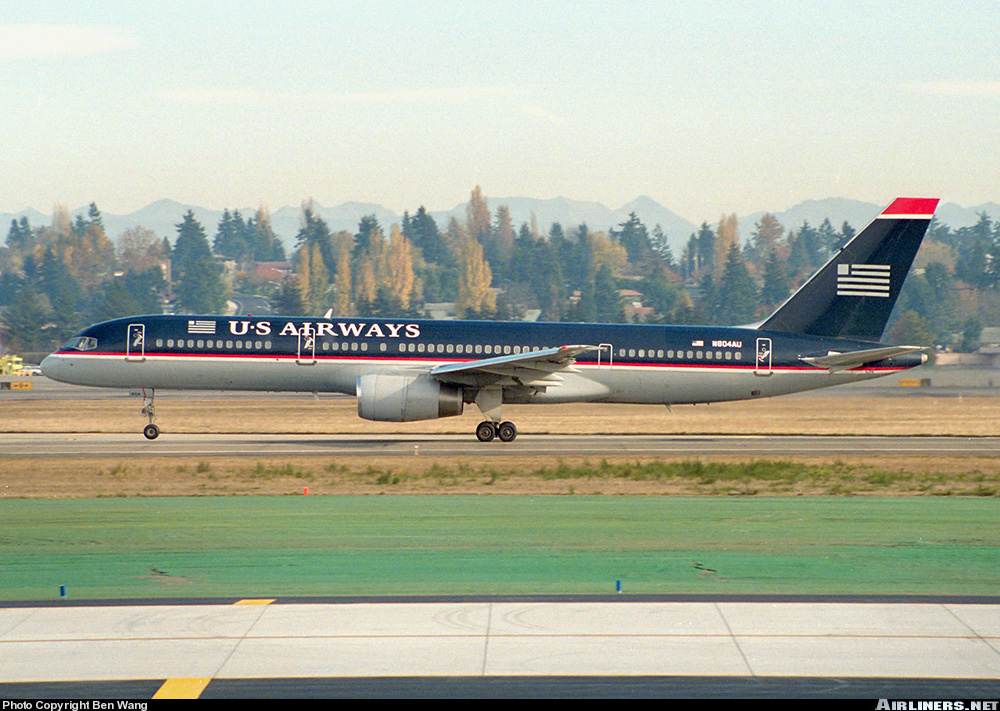}
 \includegraphics[width=0.11\textwidth, height=0.1\textwidth]{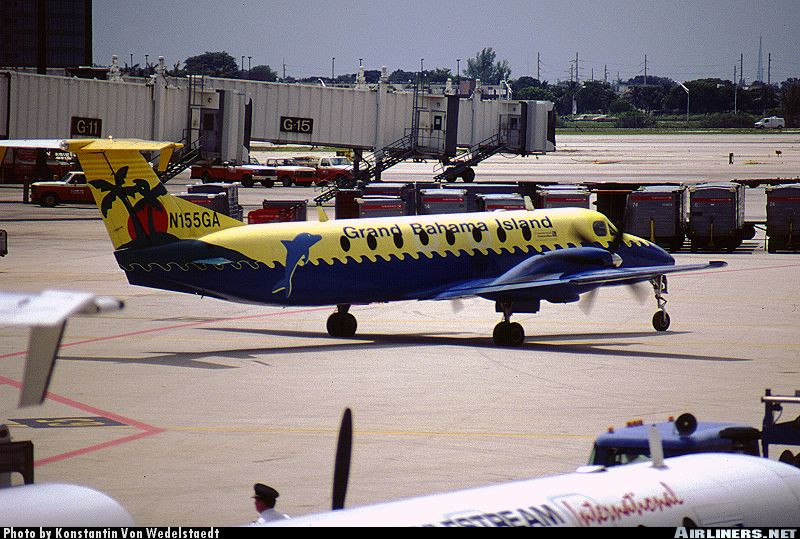}
 \includegraphics[width=0.11\textwidth, height=0.1\textwidth]{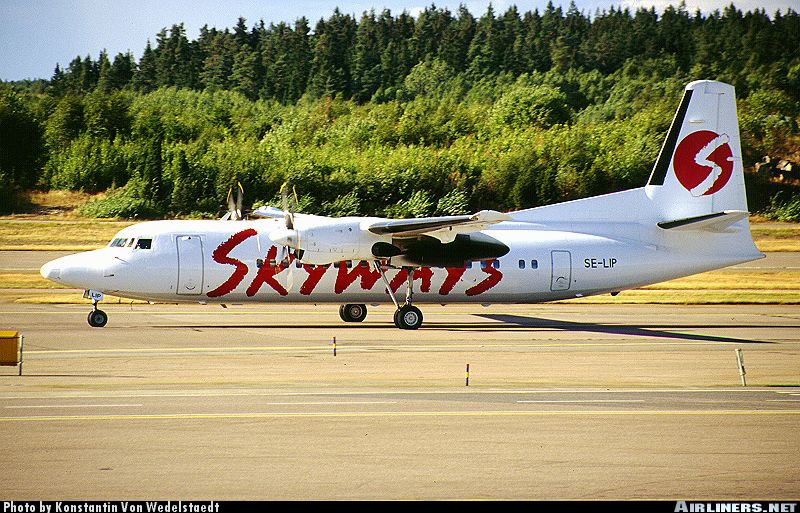}
 \includegraphics[width=0.11\textwidth, height=0.1\textwidth]{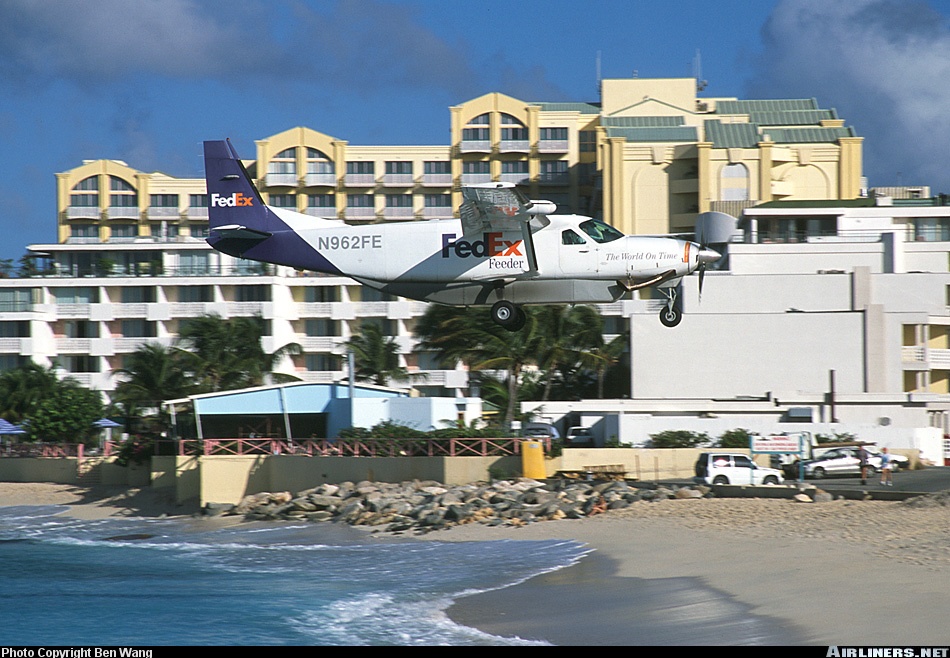}   \\
 \includegraphics[width=0.11\textwidth, height=0.1\textwidth]{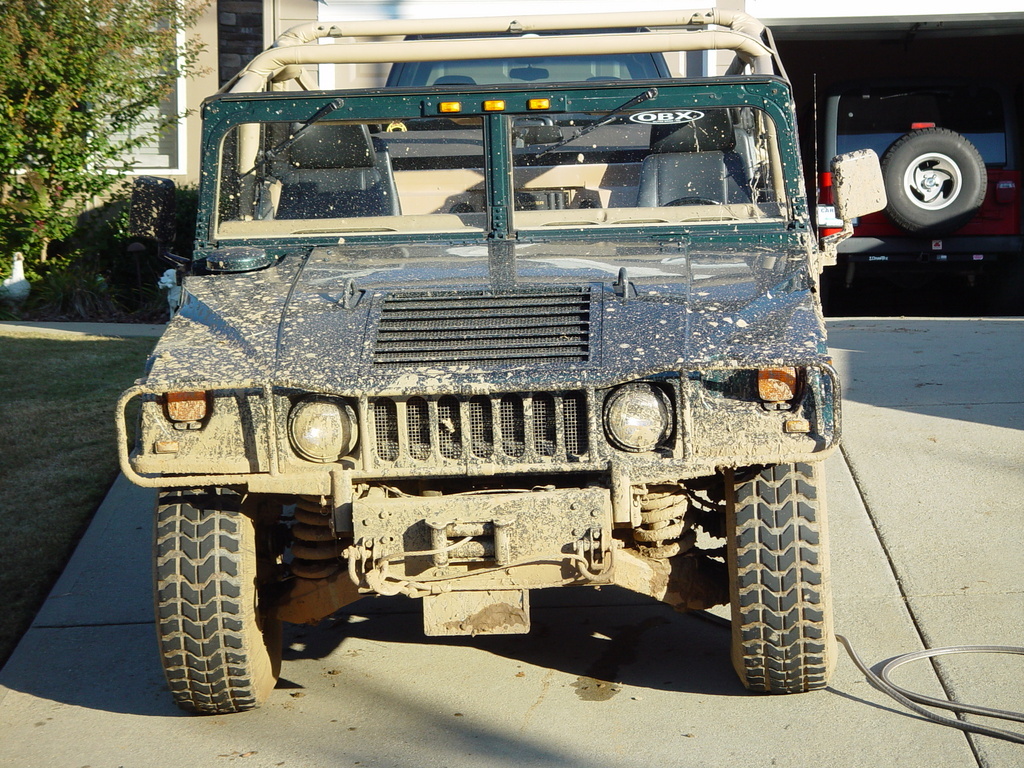}
 \includegraphics[width=0.11\textwidth, height=0.1\textwidth]{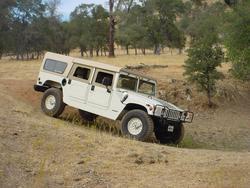}
 \includegraphics[width=0.11\textwidth, height=0.1\textwidth]{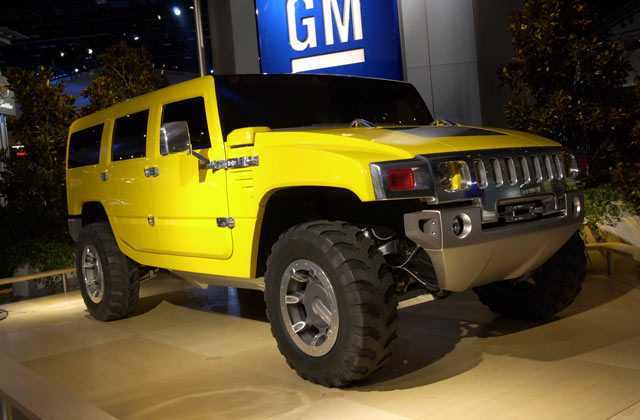}
 \includegraphics[width=0.11\textwidth, height=0.1\textwidth]{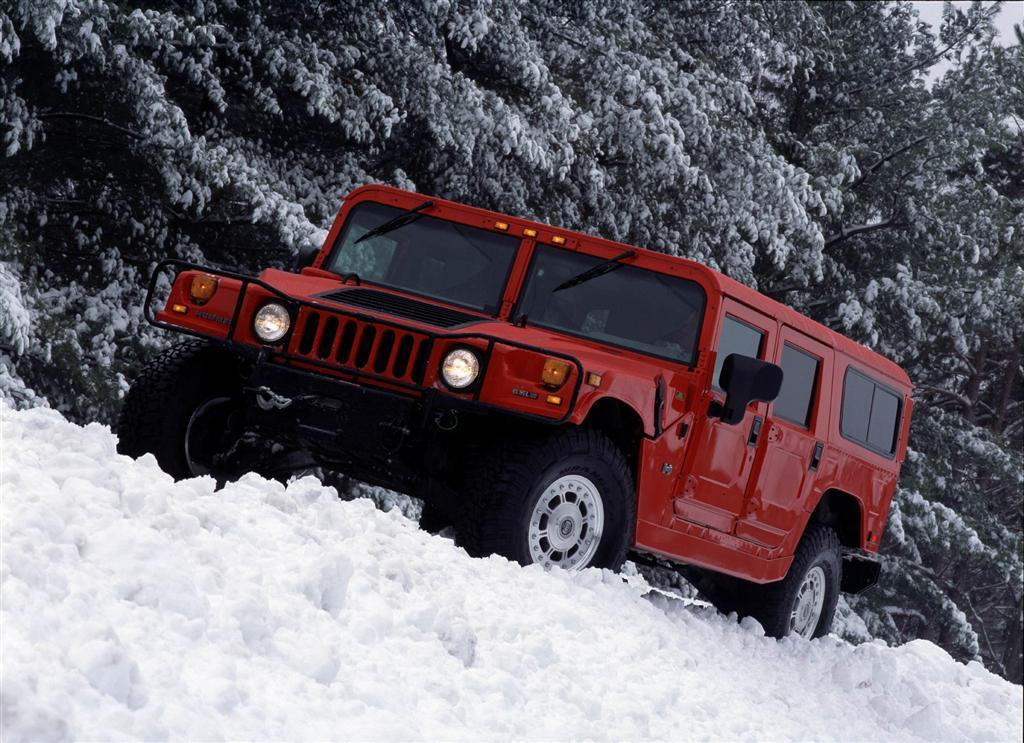}
 \includegraphics[width=0.11\textwidth, height=0.1\textwidth]{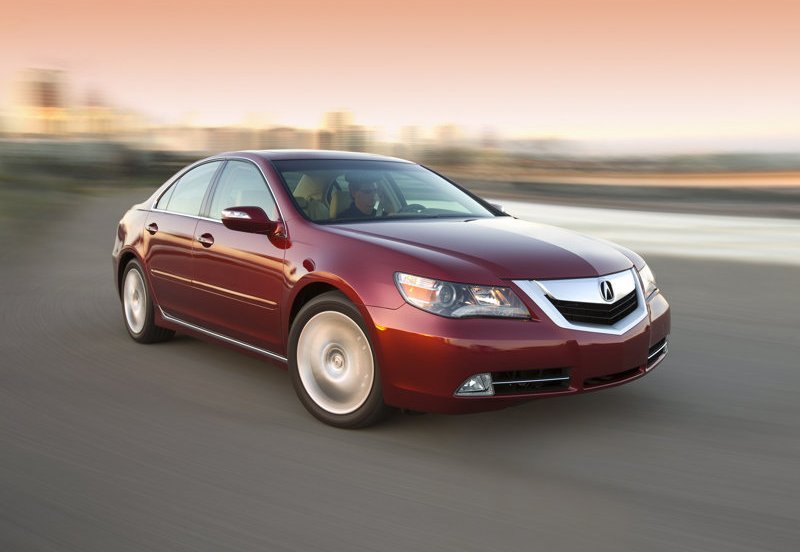}
 \includegraphics[width=0.11\textwidth, height=0.1\textwidth]{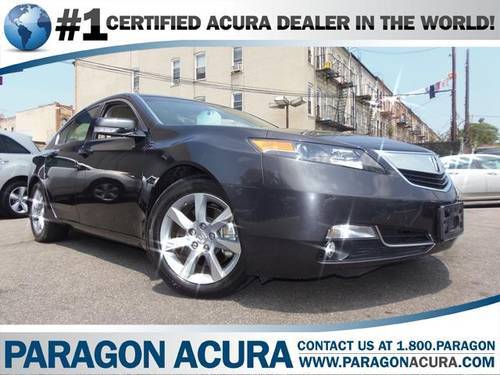}
 \includegraphics[width=0.11\textwidth, height=0.1\textwidth]{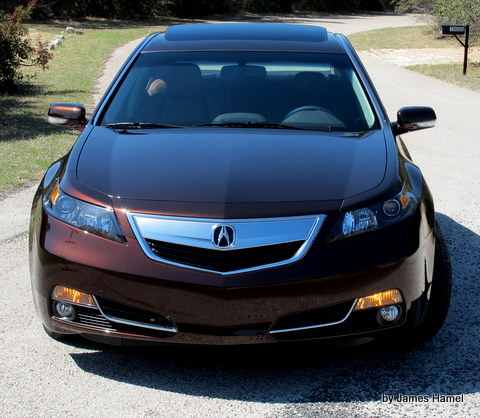}  \\
 \includegraphics[width=0.11\textwidth, height=0.1\textwidth]{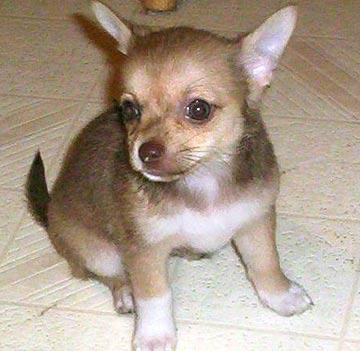}
 \includegraphics[width=0.11\textwidth, height=0.1\textwidth]{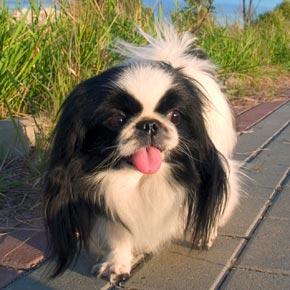}
 \includegraphics[width=0.11\textwidth, height=0.1\textwidth]{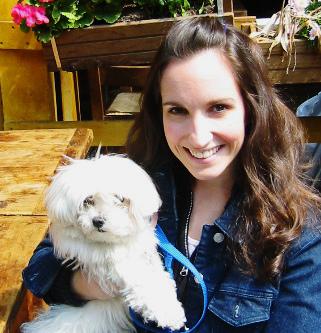}
 \includegraphics[width=0.11\textwidth, height=0.1\textwidth]{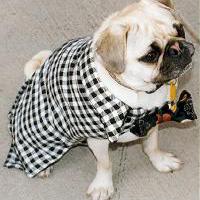}
 \includegraphics[width=0.11\textwidth, height=0.1\textwidth]{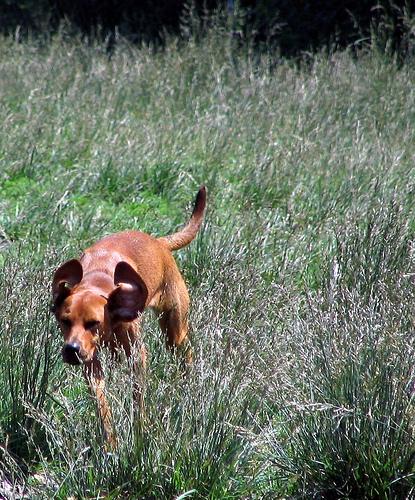}
 \includegraphics[width=0.11\textwidth, height=0.1\textwidth]{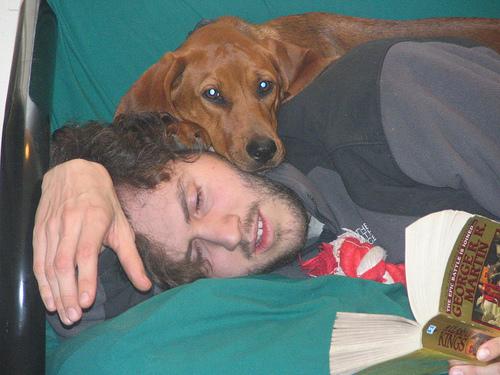}
 \includegraphics[width=0.11\textwidth, height=0.1\textwidth]{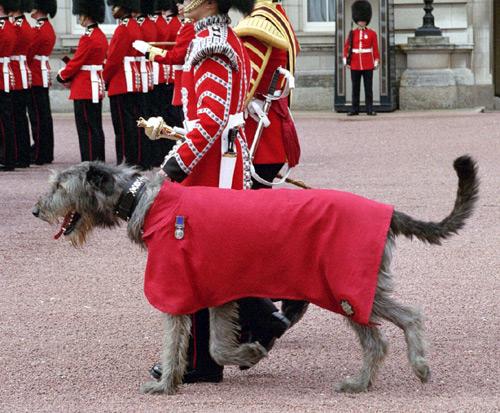}  \\
 \caption{Sample images from the datasets we use.  From top to bottom rows are images from Caltech-UCSD Birds-200-2011~\cite{WelinderEtal2010}, FGVC-Aircraft~\cite{fgvcaircraft}, cars~\cite{krause20133d} and Stanford dogs~\cite{khosla2011novel} datasets.}
 \label{fig: dataset samples}
\end{figure}

We apply our method on the famous Caltech-UCSD Birds-200-2011~\cite{WelinderEtal2010}, FGVC-Aircraft~\cite{fgvcaircraft}, cars~\cite{krause20133d} and Stanford dogs~\cite{khosla2011novel} datasets. Caltech-UCSD Birds-200-2011 dataset \cite{WelinderEtal2010} contains 11,788 images in total from 200 species of birds. Each of the 200 categories has about 60 images.  We use the default data split which randomly assigns about half of the images for each category to the training set.  FGVC-Aircraft dataset \cite{fgvcaircraft} has 100 kinds of aircrafts and contains 10,000 images in total, and we use 6,667 images for training and 3,333 for testing.  The cars~\cite{krause20133d} dataset contains 16,185 images of 196 classes of cars. The data is split into 8,144 training images and 8,041 testing images, where each class has been split roughly in a 50-50 split.  The Stanford Dogs dataset contains 20,580 images of 120 breeds of dogs from around the world and 12,000 of them are used for training, leaving 8,580 images for testing.  Some sample images from the four datasets are shown in Figure!\ref{fig: dataset samples}, where each row shows seven images from a dataset.  Our CNN-based system is implemented using Caffe~\cite{caffe}, built and tested on GoogLeNet~\cite{GoogLeNet}, VGG 19-layers~\cite{VGG} and VGG-CNN-S~\cite{chatfield2014return} in Caffe model zoo.

\subsection{Object Detection and Part Detection Results}
\label{subsubsec: obj part detection results}
\begin{figure}[h!!]
 \centering
 \includegraphics[width=0.95\textwidth]{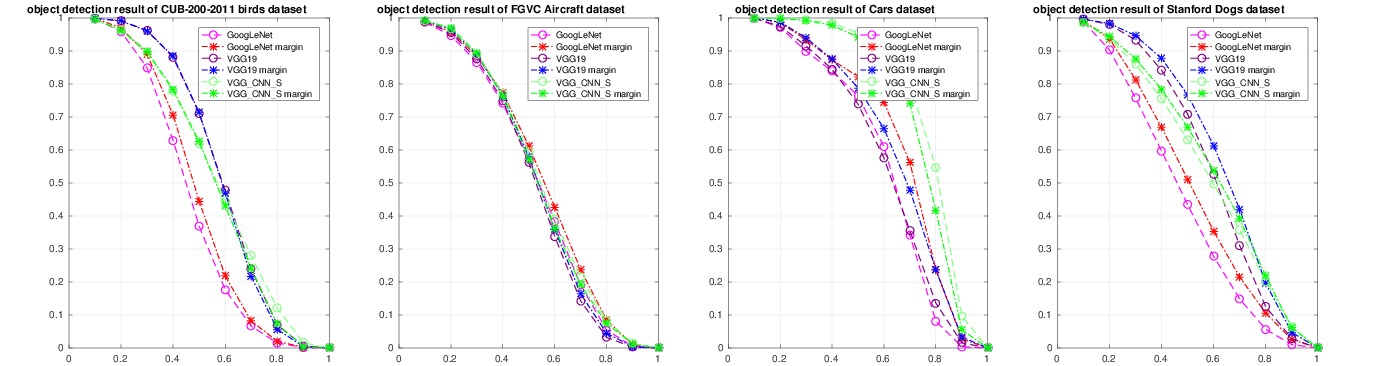}
 \caption{Recall vs. IoU of our CNN-based automatic object detection results.  From left to right are the plots of Caltech-UCSD Birds-200-2011~\cite{WelinderEtal2010}, FGVC-Aircraft~\cite{fgvcaircraft}, cars~\cite{krause20133d} and Stanford dogs~\cite{khosla2011novel} datasets respectively.  Each plot contains the results based on three CNN models: GoogLeNet~\cite{GoogLeNet}, VGG 19-layers~\cite{VGG} and VGG-CNN-S~\cite{chatfield2014return}.}
 \label{fig: recall to IoU}
\end{figure}

Our algorithm detects the object by summing the feature maps of layers in $\mathcal{S}^{layer\_idx}_{object}$, and the bounding box can be obtained by directly cropping the rectangle that tightly contains the thresholded active region.  However, to be conservative, we extend the height and width of the rectangle by $5\%$.  Both the results with and without conservative extension are evaluated using Recall-to-IoU shown in Figure~\ref{fig: recall to IoU} (better viewed in color).  The results with extended boundary use star markers and are annotated by `margin' in the legend.  From left to right are the plots of Caltech-UCSD Birds-200-2011~\cite{WelinderEtal2010}, FGVC-Aircraft~\cite{fgvcaircraft}, cars~\cite{krause20133d} and Stanford dogs~\cite{khosla2011novel} datasets respectively.  Each plot contains the results based on three CNN models: GoogLeNet~\cite{GoogLeNet}, VGG 19-layers~\cite{VGG} and VGG-CNN-S~\cite{chatfield2014return}.  It can be seen that the detection results vary between network models and datasets. In general, relatively shallower networks like VGGs~\cite{VGG,chatfield2014return} outperforms the deep GoogLeNet~\cite{GoogLeNet}, since pooling layers reduce the resolution of the feature maps.  The results with conservative extension are slightly better than that without.

\begin{figure}[h!!]
 \centering
 \includegraphics[width=0.98\textwidth]{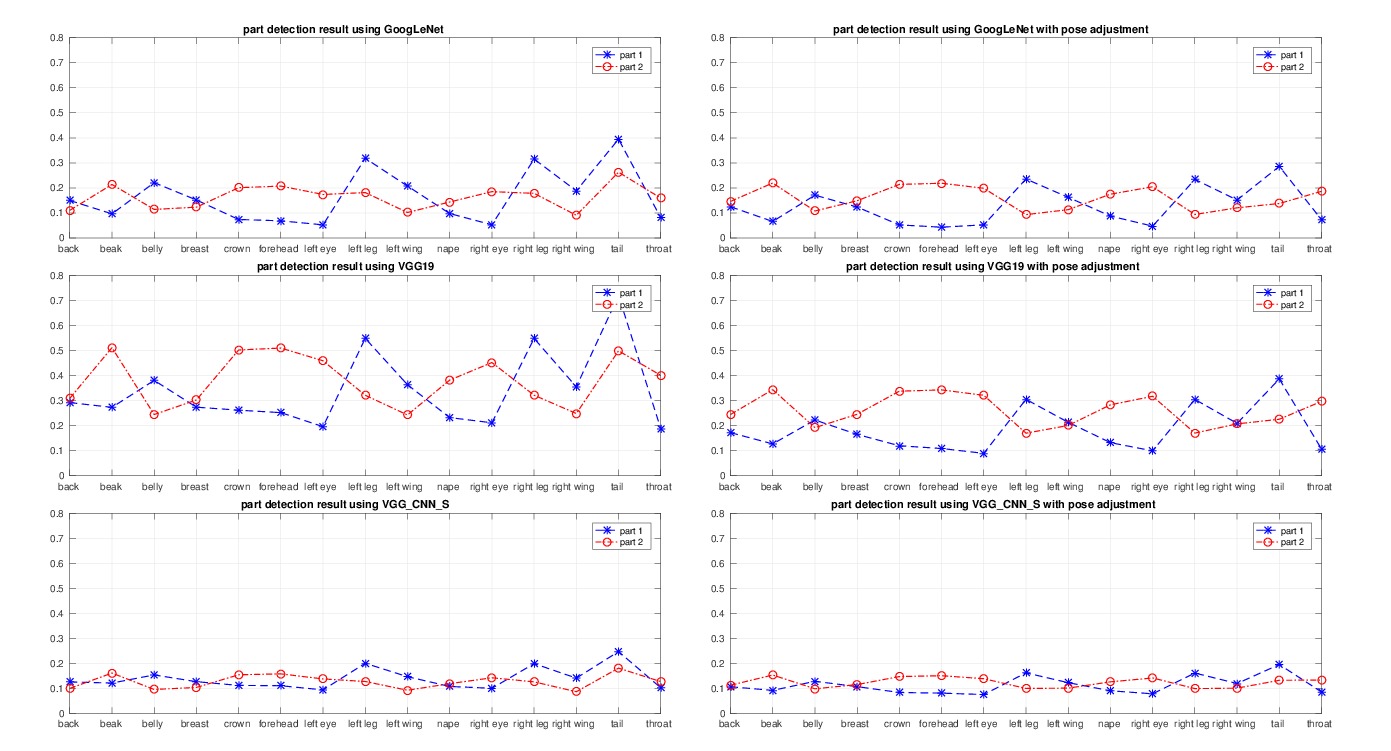}
 \caption{Part detection results of Caltech-UCSD Birds-200-2011~\cite{WelinderEtal2010}.  (The names and ID numbers of the parts are the same as listed in Table~\ref{tab:CUB parts})}
 \label{fig: part detection result}
\end{figure}

Among the four datasets only Caltech-UCSD Birds-200-2011~\cite{WelinderEtal2010} offers the ground truth part annotations.  The locations of 15 parts listed in Table~\ref{tab:CUB parts} are specified by 2-D coordinates if visible.  Our system detects two parts of the bird.  For each detected part, we calculate the difference between its coordinates to all the available annotated parts.  Next ,we first normalize the x and y coordinates of the difference by the width and height of the ground truth bounding box respectively, then calculate its norm.  The averaged results are shown in Figure~\ref{fig: part detection result}.  The left column in Figure~\ref{fig: part detection result} shows the detection results directly obtained from the activation of the CNN hidden layer feature maps, the right column shows the results after shifting based on the pose estimated in Subsec~\ref{subsubsec: part-pose}.  The three plots in each column from top to bottom show the detection results using GoogLeNet~\cite{GoogLeNet}, VGG 19-layers~\cite{VGG} and VGG-CNN-S~\cite{chatfield2014return} respectively.  It can be seen that both the curves of part 1 (i.e. the head) and part 2 (i.e. the body) have their featured shapes which are consistently kept through all the cases. Part 1 is always very close to the eyes, the forehead, the throat and the crown, while very far away from the belly, the wings, the legs and the tail.  On the contrary, part 2 behaves almost oppositely.  In the left column, both part 1 and part 2 are relatively far away from the tail and part 1 is much farther than part 2.  After shifting based on the pose, the center of part 2 move much closer to the tail.   In general, the cures in the right column are lower than that in the left column, which means that to adjust the detection results based on the estimated pose increases the part detection accuracy.

\begin{table}[h]
\begin{center}
{\small
\begin{tabular}{|l||*{8}{@{}c@{}|}}
\hline
\multicolumn{1}{|c||}{ID}  & 1 & 2 & 3 & 4 & 5 & 6 & 7 & 8  \\
\hline
name           & back & { beak } & belly & breast & crown & { forehead } & { left eye } & { left leg } \\
\hline\hline
\multicolumn{1}{|c||}{ID}   & 9 & 10 & 11 & 12 & 13 & 14 & 15 &  \\
\hline
name           & { left wing } & nape  & { right eye } & { right leg } & { right wing } & tail & throat &   \\    
\hline
\end{tabular}
}
\end{center}
\caption{The 15 parts annotated with ground truth location coordinates in Caltech-UCSD Birds-200-2011~\cite{WelinderEtal2010}.}
\label{tab:CUB parts}
\end{table}

\subsection{Robustness of Detection Results to $T_{object}$ and $T_{parts}$}
\label{subsubsec: robustness of detection results}
\begin{figure}[h!]
 \centering
 \includegraphics[width=0.9\textwidth, height=0.1\textwidth]{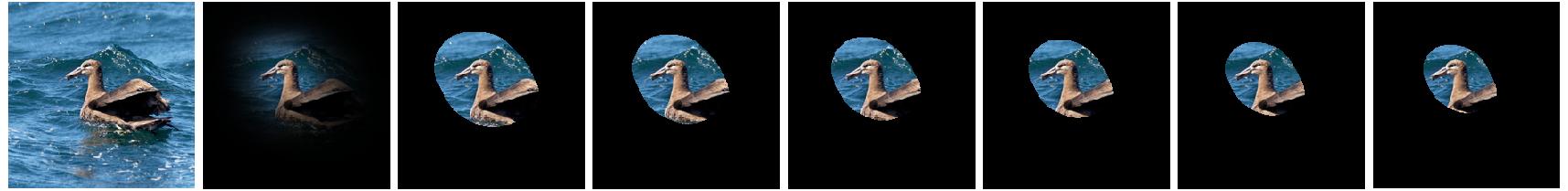}
 \includegraphics[width=0.9\textwidth, height=0.1\textwidth]{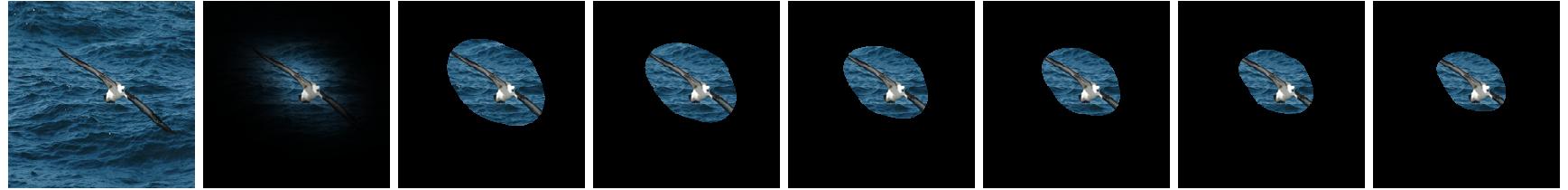}
 \includegraphics[width=0.9\textwidth, height=0.1\textwidth]{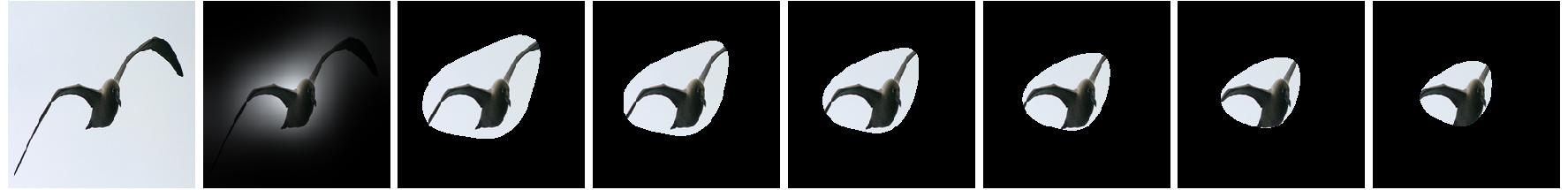}
 \includegraphics[width=0.9\textwidth, height=0.1\textwidth]{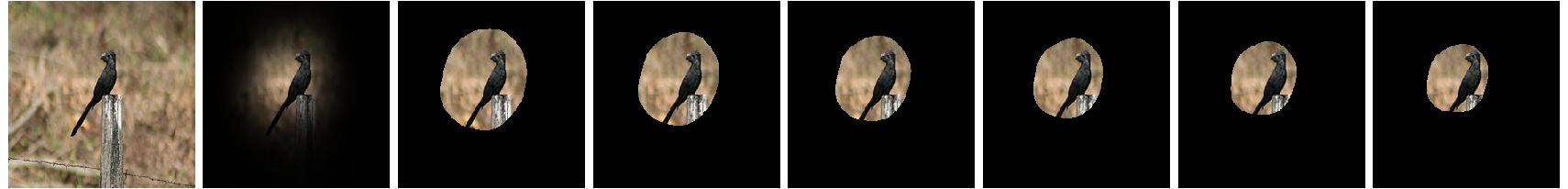}
 \includegraphics[width=0.9\textwidth, height=0.1\textwidth]{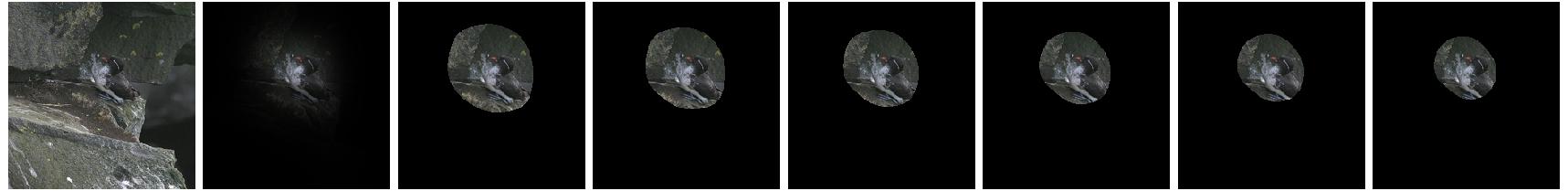}
 \includegraphics[width=0.9\textwidth, height=0.1\textwidth]{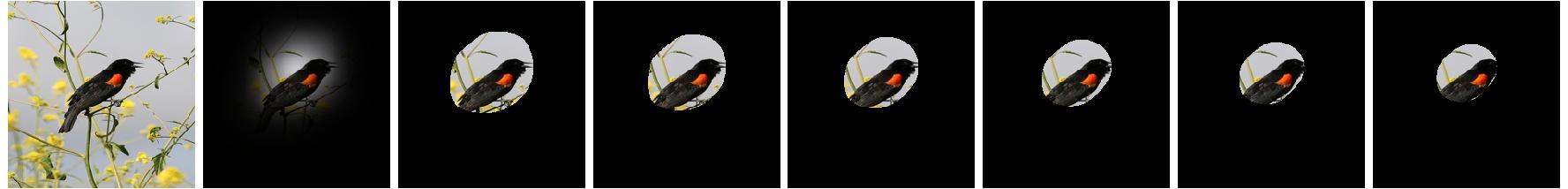}
 \includegraphics[width=0.9\textwidth, height=0.1\textwidth]{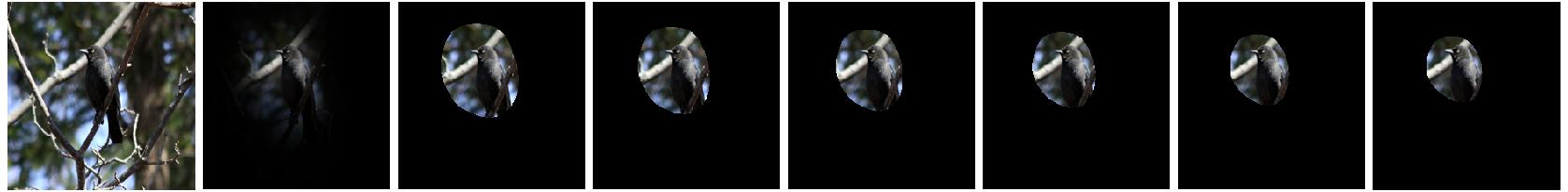}
 \includegraphics[width=0.9\textwidth, height=0.1\textwidth]{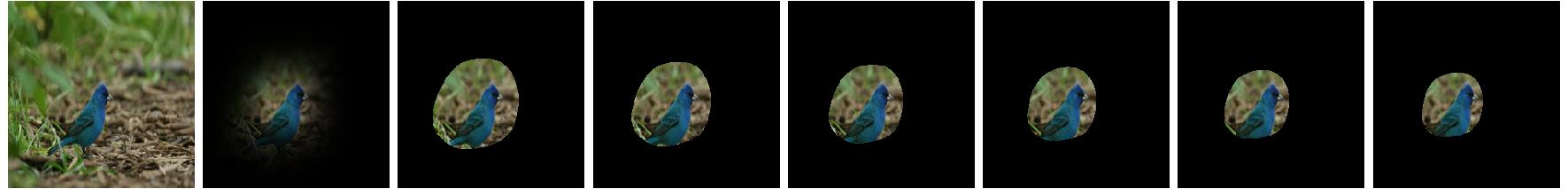} 
 \includegraphics[width=0.9\textwidth, height=0.1\textwidth]{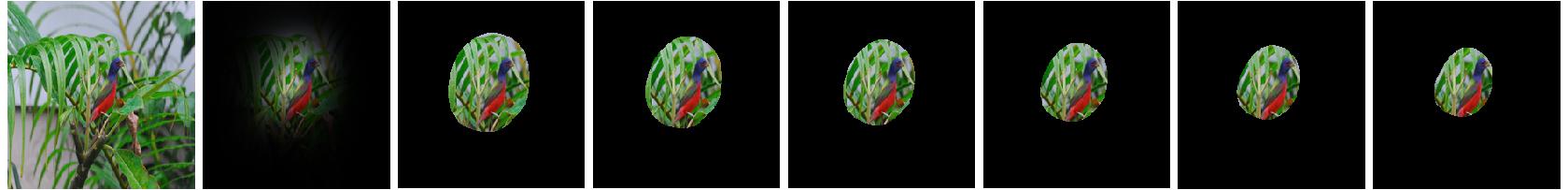}
 \caption{Illustration of object detection with $T_{object}=0.2,0.25,0.3,0.35,0.4,0.45$.}
 \label{fig:obj threshold}
\end{figure}

\begin{figure}[h!]
 \centering
 \includegraphics[width=0.9\textwidth, height=0.2\textwidth]{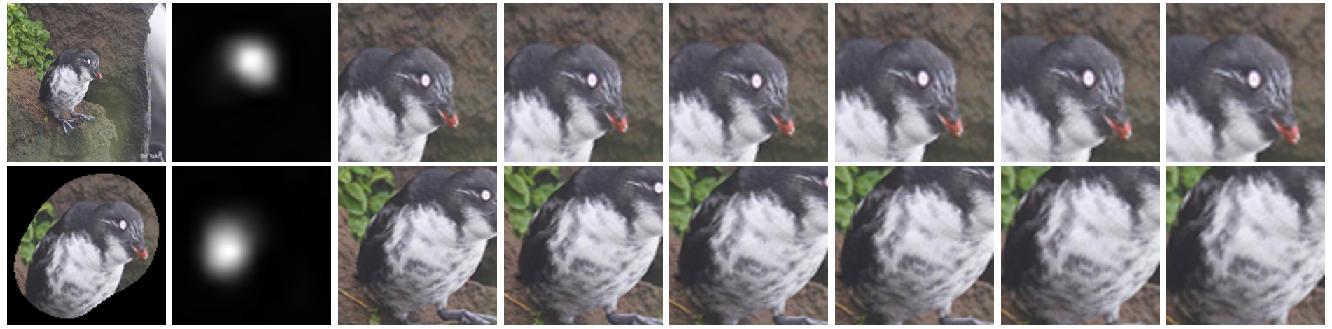}
 \includegraphics[width=0.9\textwidth, height=0.2\textwidth]{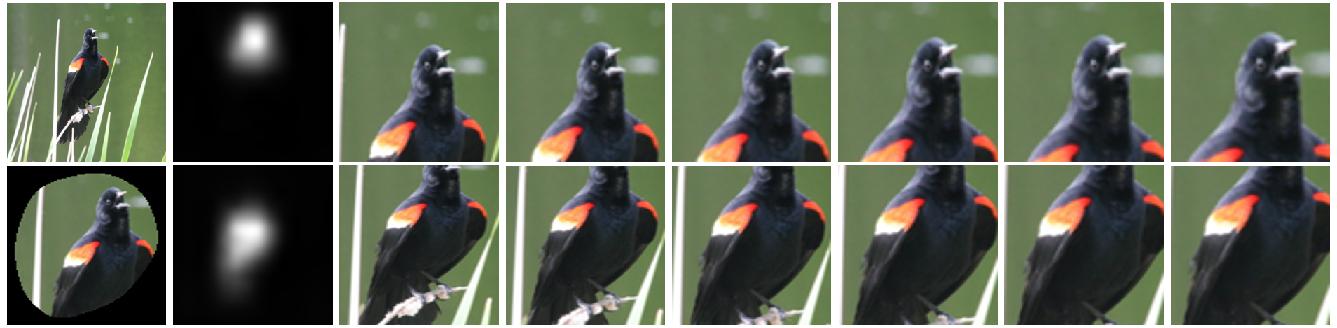}
 \includegraphics[width=0.9\textwidth, height=0.2\textwidth]{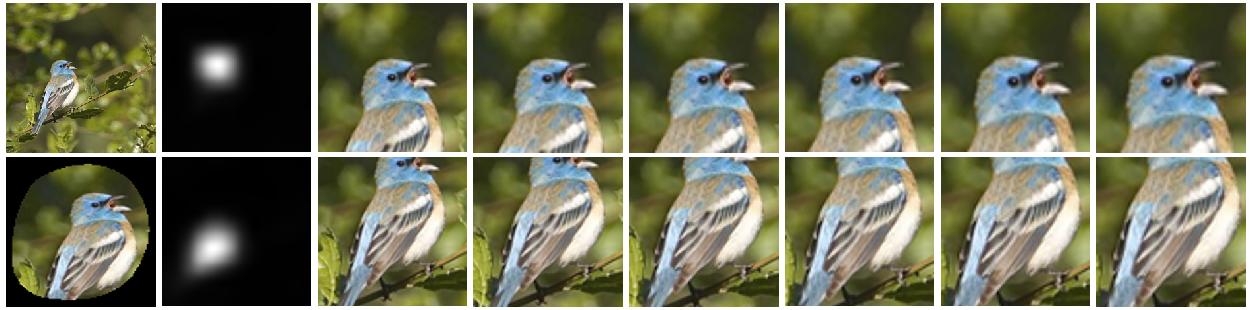}
 \includegraphics[width=0.9\textwidth, height=0.2\textwidth]{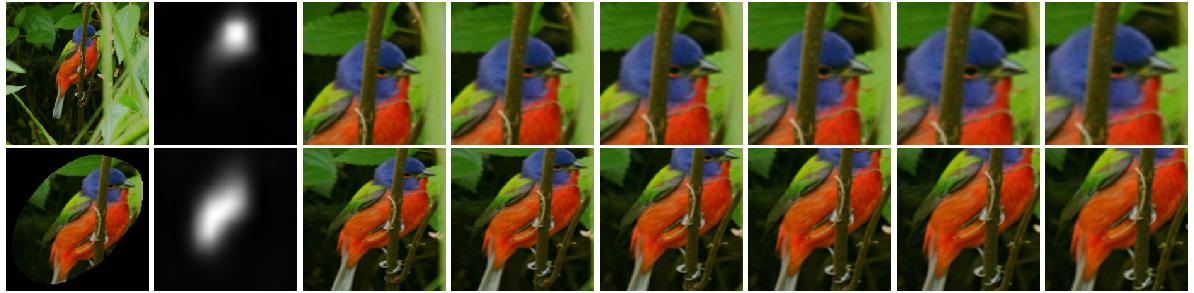} 
 \includegraphics[width=0.9\textwidth, height=0.2\textwidth]{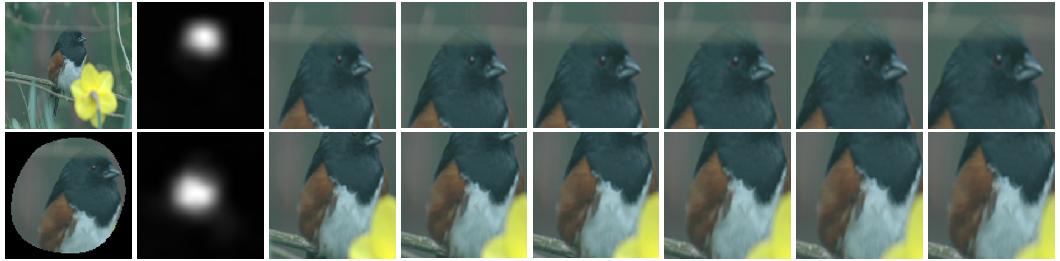}
 \includegraphics[width=0.9\textwidth, height=0.2\textwidth]{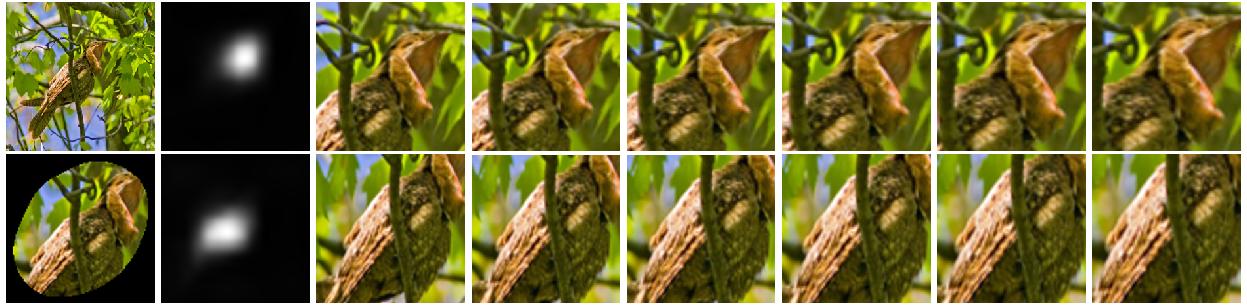}
 \includegraphics[width=0.9\textwidth, height=0.2\textwidth]{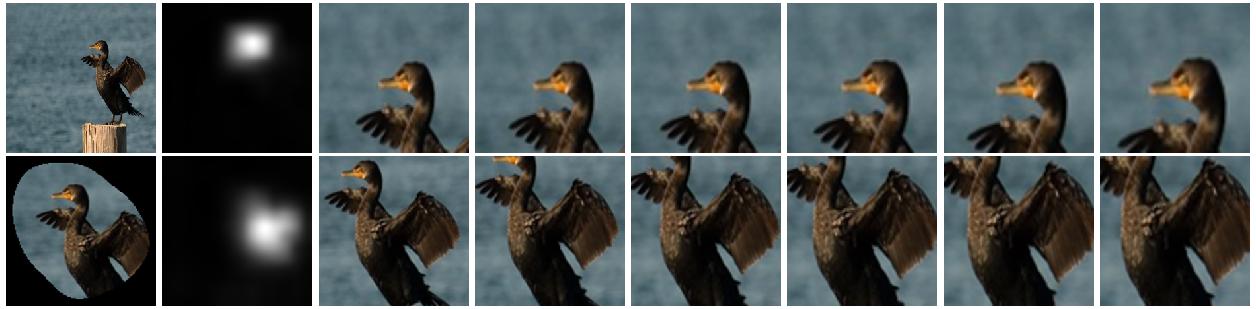}
 \caption{Illustration of part detection with $T_{parts}=0.2,0.25,0.3,0.35,0.4,0.45$.}
 \label{fig:part threshold}
\end{figure}

Figure~\ref{fig:obj threshold} illustrates our object detection result and how it varies with $6$ different choices of $T_{object}$. The first two images in each row show a resized original image and its soft-masked version. We threshold the soft mask $\mathbf{M}_{object}$ with $T_{object}=0.2,0.25,\cdots ,0.45$ to obtain $\mathbf{M}^{binary}_{object}$ and display the result of applying $\mathbf{M}^{binary}_{object}$ to the original images. It can be seen that, as expected, the higher $T_{object}$ is, the smaller is the active region in $\mathbf{M}^{binary}_{object}$.  In general, the object detection result is not sensitive to $T_{object}$, which means that the object saliency masks are rather confident to distinguish between the foreground and background.  Similar observation can be found in Figure~\ref{fig:part threshold} for part detection of $N_{part} = 2$, in which each example is shown in two consecutive rows.  The original image is displayed in the upper-left corner and below it shows the object-focused image masked by $\mathbf{M}^{binary}_{object}$. Column $2$ shows the masks of two detected parts with the one having a smaller region on top. Columns $3$ to $8$ show the part-focused images generated with $T_{parts}=0.2,0.25,\cdots ,0.45$. To be conservative, when we crop the object-focused and part-focused images, we extend the height and width of the rectangle by $5\%$. It is clear that the two parts focused are the head and body of a bird and this result is quite robust.

\subsection{Explain away Ensemble}
\label{subsec:explain away ensemble}
Though our method integrates multiple CNNs, the performance gain does not come from ensemble.  This is proofed with the classification results in Table~\ref{tab:CUB200 result} and Table~\ref{tab:FGVC Aircraft result}, which include our base line, the result using proposed method with $N_{part} = 1, 2, 3, 4$, and the ensemble results using the same network structure.  `GoogLeNet-ft' refers to the fine-tuned GoogLeNet~\cite{GoogLeNet} pre-trained on the ImageNet dataset \cite{ILSVRC15}, with its classification layer replaced to suit our fine-grained categorization datasets; `2 CNNs ensemble' means that the input of the two integrated CNNs are both the raw images, similarly for `3 CNNs ensemble' and `4 CNNs ensemble'; `2 CNNs: raw + obj' means using the concatenated feature of two CNNs, one input is original raw images, the other input is object-focused images ($N_{part}=1$); `3 CNNs: raw+2 parts' means the inputs of the 3 integrated CNNs are raw images and 2 parts-focused images, similarly for `4 CNNs: raw+3 parts'.  Comparing the results of `GoogLeNet-ft' and `3 CNNs: raw+2 parts', it can be seen that the proposed method greatly outperforms our baseline ($81.38\%$ vs. $73\%$, $82.7\%$ vs. $75\%$).  With the same number of CNNs integrated, the performance achieved by simple `ensemble' method largely lags behind the proposed method.  This shows the effectiveness of our CNN-based part feature extraction on fine-grained classification.  From the results of using different $N_{part}$, we can see that the best performance is achieved by $N_{part} = 2$,  which is consistent with our analysis in Subsec~\ref{subsubsec: set N part}.

\begin{table}[h]
\begin{center}
{\small
\begin{tabular}{|p{1.4cm}|p{1.4cm}|p{1cm}|p{1cm}|p{1cm}|p{1cm}|p{1cm}|p{1cm}|} 
\hline
\multicolumn{1}{|c|}{Method}  &  GoogLeNet-ft (baseline)  & 2 CNNs ensemble & 2 CNNs: raw+obj & 3 CNNs ensemble & 3 CNNs: raw+2 parts & 4 CNNs ensemble & 4 CNNs: raw+3 parts \\
\hline\hline
Test Accuracy ($\%$)      &  $73$    & $72.7$    & $77$     & $74$    & $81.38$  &  $72.1$  & $78.5$  \\
\hline
\end{tabular}
}
\end{center}
\caption{Fine-grained categorization results on Caltech-UCSD Birds-200-2011 dataset of using proposed system with various settings. Detailed description can be found in the text.}
\label{tab:CUB200 result}
\vspace{-1em}
\end{table}

\begin{table}[h]
\begin{center}
{\small
\begin{tabular}{|p{1.4cm}|p{1.4cm}|p{1cm}|p{1cm}|p{1cm}|p{1cm}|p{1cm}|p{1cm}|} 
\hline
\multicolumn{1}{|c|}{Method}  &  GoogLeNet-ft (baseline) & 2 CNNs ensemble & 2 CNNs: raw+obj & 3 CNNs ensemble & 3 CNNs: raw+2 parts & 4 CNNs ensemble & 4 CNNs: raw+3 parts \\
\hline\hline
Test Accuracy ($\%$)      &  $75$    & $77.0$   & $80$    & $78.57$     & $82.7$    &  $76.6$  & $80.1$  \\
\hline
\end{tabular}
}
\end{center}
\caption{Fine-grained categorization results on FGVC Aircraft dataset of using proposed system with various settings. Detailed description can be found in the text.}
\label{tab:FGVC Aircraft result}
\vspace{-1em}
\end{table}

\subsection{Comparison with Other Methods}
\label{subsec:comparison with other methods}
Table~\ref{tab:CUB result1}, Table~\ref{tab:FGVC result}, Table~\ref{tab:CUB result2}, Table~\ref{tab:cars result} and Table~\ref{tab:dogs result} compares many existing methods applied to the Caltech-UCSD Birds-200-2011~\cite{WelinderEtal2010}, FGVC-Aircraft~\cite{fgvcaircraft}, cars~\cite{krause20133d} and Stanford dogs~\cite{khosla2011novel} datasets.  The proposed method is built and tested on GoogLeNet~\cite{GoogLeNet}, VGG 19-layers~\cite{VGG} and VGG-CNN-S~\cite{chatfield2014return}.  Directly fine-tuning these CNNs can be viewed as our baselines.  The result of proposed method together with their baseline results (in total 6 terms) are listed at the bottom.  For all our fine-tuning operations, we adopt the two-step fine-tuning method in~\cite{birdspecies}.  Similar as \cite{AlexNet}, during testing, we crop five images from the original image at the four corners and the center, flipping them to form 10 crops.  Doing so leads to an increase in accuracy by about $1\%$.  We can see that our proposed method improves the performance of our baselines in all cases with maximum categorization accuracy increased by $14.26\%$ (using VGG-CNN-S~\cite{chatfield2014return} on cars~\cite{krause20133d}), and achieves superior or comparable performance with that of the state-of-the-art under the general problem setting that only class labels are available during training with no other annotations.  Importantly, instead of assembling complex algorithms, our proposed method only takes advantage of the CNN feature maps themselves to achieve good though not the best performance. 

\begin{table}[h]
\begin{center}
{\small
\begin{tabular}{|l|c c|c|}
\hline
\multicolumn{1}{|c|}{Method}            &  Train Anno  & Test Anno &  Accuracy ($\%$) \\
\hline\hline
POOF \cite{POOF}                        &  $B+P$       & $B+P$     & $73.3\ \ $   \\
PN-CNN \cite{birdspecies}               &  $B+P$       & $B+P$     & $85.4\ \ $   \\
Symbiotic \cite{symbiotic}              &  $B+P$       & $B+P$     & $69.5\ \ $  \\
\hline
Part R-CNN \cite{zhang2014part}         &  $B+P$       & $B$       & $76.37$  \\
POOF \cite{POOF}                        &  $B+P$       & $B$       & $56.8\ \ $   \\
DPD+DeCAF \cite{decaf}                  &  $B+P$       & $B$       & $64.96$  \\
Nonparametric \cite{nonparametric}      &  $B+P$       & $B$       & $57.84$  \\
Symbiotic \cite{symbiotic}              &  $B+P$       & $B$       & $61.60$  \\
Deep LAC \cite{lin2015deep}             &  $B+P$       & $B$       & $80.26$  \\
Spda-cnn \cite{zhang2016spda}           &  $B+P$       & $B$       & $85.14$  \\
PBC \cite{huang2016pbc}                 &  $B+P$       & $B$       & $83.7\ \ $  \\
\hline
No parts \cite{krause2015fine}          &  $B$         & $B$       & $82.8\ \ $   \\
Symbiotic \cite{symbiotic}              &  $B$         & $B$       & $59.4\ \ $   \\
CNNaug-SVM \cite{razavian2014cnn}       &  $B$         & $B$       & $61.8\ \ $   \\
Bilinear \cite{bilinear}                &  $B$         & $B$       & $85.1\ \ $   \\
multi-stage \cite{qian2015fine}         &  $B$         & $B$       & $67.86$   \\
Coarse-to-fine \cite{yao2016coarse}     &  $B$         & $B$       & $82.9\ \ $   \\
Simple tech \cite{xie2016simple}        &  $B$         & $B$       & $66.87 $   \\
FOAF  \cite{zhang2016fused}             &  $B$         & $B$       & $86.34 $   \\
\hline
PS-CNN \cite{huang2016part}             &  $B+P$       & $B$       & $76.6\ \ $   \\
\hline
PN-CNN \cite{birdspecies}               &  $B+P$       & $-$       & $75.7\ \ $   \\
Part R-CNN \cite{zhang2014part}         &  $B+P$       & $-$       & $73.89$  \\
\hline
No parts \cite{krause2015fine}          &  $B$         & $-$       & $82.0\ \ $   \\
Coarse-to-fine \cite{yao2016coarse}     &  $B$         & $-$       & $82.5\ \ $   \\
Mul-granularity \cite{wang2015multiple} &  $B$         & $-$       & $83.0\ \ $  \\
Task-driven \cite{huang2016task}        &  $B$         & $-$       & $81.69 $   \\
PBC \cite{huang2016pbc}                 &  $B$         & $-$       & $83.3\ \ $  \\
\hline
\end{tabular}
}
\end{center}
\caption{Comparison of different classification methods on Caltech-UCSD Birds-200-2011 dataset \cite{WelinderEtal2010}. `Anno' stands for `annotations', `B' for `bounding box', `P' for `part location' and `ft' for `fine-tune'. The rounding precision of other methods may differ a little in other citations.}
\label{tab:CUB result1}
\end{table}

\begin{table}[h]
\begin{center}
{\small
\begin{tabular}{|l|c c|c|}
\hline
\multicolumn{1}{|c|}{Method}            &  Train Anno  & Test Anno &  Accuracy ($\%$) \\
\hline\hline
Two-level \cite{xiao2014application}    &  $-$         & $-$       & $69.7\ \ $   \\
Constellation \cite{simon2015neural}    &  $-$         & $-$       & $81.01$  \\
Spatial trans \cite{simon2015neural}    &  $-$         & $-$       & $84.1\ \ $  \\
W-supervised \cite{weakly}              &  $-$         & $-$       & $79.34 $   \\
Bilinear \cite{bilinear}                &  $-$         & $-$       & $84.1\ \ $  \\
Mul-granularity \cite{wang2015multiple} &  $-$         & $-$       & $81.7\ \ $  \\
Picking deep \cite{zhang2016picking}    &  $-$         & $-$       & $84.54   $  \\
Annotation-modi \cite{luo2016annotation}&  $-$         & $-$       & $75.36 $  \\
Fused \cite{zhang2016fused}             &  $-$         & $-$       & $84.63 $   \\
Friend-or-Foe \cite{xu2017friend}       &  $-$         & $-$       & $77.4\ \ $  \\
GoogLeNet-ft                            &  $-$         & $-$       & $73.0\ \ $  \\
Ours-GoogLeNet-based                    &  $-$         & $-$       & $81.38$   \\
VGG-19-ft                               &  $-$         & $-$       & $70.79$   \\
Ours-VGG-19-based                       &  $-$         & $-$       & $72.0\ \ $   \\
VGG-CNN-S-ft                            &  $-$         & $-$       & $56.21$   \\
Ours-VGG-CNN-S-based                    &  $-$         & $-$       & $58.83 $   \\
\hline
\end{tabular}
}
\end{center}
\caption{Comparison of different classification methods on Caltech-UCSD Birds-200-2011 dataset \cite{WelinderEtal2010}. It is continued from Table~\ref{tab:CUB result1}. The rounding precision of other methods may differ a little in other citations.}
\label{tab:CUB result2}
\end{table}

\begin{table}[h]
\begin{center}
{\small
\begin{tabular}{|l|c c|c|}
\hline
\multicolumn{1}{|c|}{Method}            &  Train Anno  & Test Anno &  Accuracy ($\%$) \\
\hline\hline
Coarse-to-fine \cite{yao2016coarse}     &  $B$         & $B$       & $87.7\ \ $   \\
Simple tech \cite{xie2016simple}        &  $B$         & $B$       & $72.18 $   \\
Mining Triplets \cite{wang2016mining}   &  $B$         & $B$       & $88.4\ \ $   \\
\hline
Coarse-to-fine \cite{yao2016coarse}     &  $B$         & $-$       & $86.9\ \ $   \\
Mul-granularity \cite{wang2015multiple} &  $B$         & $-$       & $86.6\ \ $  \\
\hline
Bilinear \cite{bilinear}                &  $-$         & $-$       & $84.1\ \ $  \\
Mul-granularity \cite{wang2015multiple} &  $-$         & $-$       & $82.5\ \ $  \\
GoogLeNet-ft                            &  $-$         & $-$       & $75.0\ \ $  \\
Ours-GoogLeNet-based                    &  $-$         & $-$       & $82.7\ \ $   \\
VGG-19-ft                               &  $-$         & $-$       & $77.88$   \\
Ours-VGG-19-based                       &  $-$         & $-$       & $81.8\ \ $   \\
VGG-CNN-S-ft                            &  $-$         & $-$       & $66.48$   \\
Ours-VGG-CNN-S-based                    &  $-$         & $-$       & $69.5\ \ $   \\
\hline
\end{tabular}
}
\end{center}
\caption{Comparison of different classification methods on FGVC-Aircraft dataset~\cite{fgvcaircraft}. `Anno' stands for `annotations', `B' for `bounding box', `P' for `part location' and `ft' for `fine-tune'. The rounding precision of other methods may differ a little in other citations.}
\label{tab:FGVC result}
\end{table}

\begin{table}[h]
\begin{center}
{\small
\begin{tabular}{|l|c c|c|}
\hline
\multicolumn{1}{|c|}{Method}          &  Train Anno  & Test Anno &  Accuracy ($\%$) \\
\hline\hline
No parts \cite{krause2015fine}        &  $B$         & $B$       & $92.8\ \ $   \\
Mining Triplets \cite{wang2016mining} &  $B$         & $B$       & $92.5\ \ $   \\
\hline
Learn parts \cite{krause2014learning} &  $B$         & $-$       & $73.9\ \ $   \\
No parts \cite{krause2015fine}        &  $B$         & $-$       & $92.6\ \ $   \\
\hline
Bilinear \cite{bilinear}              &  $-$         & $-$       & $91.3\ \ $  \\
GoogLeNet-ft                          &  $-$         & $-$       & $81.84 $  \\
Ours-GoogLeNet-based                  &  $-$         & $-$       & $85.42$   \\
VGG-19-ft                             &  $-$         & $-$       & $78.74$   \\
Ours-VGG-19-based                     &  $-$         & $-$       & $83.0\ \ $   \\
VGG-CNN-S-ft                          &  $-$         & $-$       & $60.49$   \\
Ours-VGG-CNN-S-based                  &  $-$         & $-$       & $74.75$   \\
\hline
\end{tabular}
}
\end{center}
\caption{Comparison of different classification methods on cars dataset~\cite{krause20133d}. `Anno' stands for `annotations', `B' for `bounding box', `P' for `part location' and `ft' for `fine-tune'. The rounding precision of other methods may differ a little in other citations.}
\label{tab:cars result}
\end{table}

\begin{table}[h]
\begin{center}
{\small
\begin{tabular}{|l|c c|c|}
\hline
\multicolumn{1}{|c|}{Method}          &  Train Anno  & Test Anno &  Accuracy ($\%$) \\
\hline\hline
Symbiotic \cite{symbiotic}            &  $B$         & $B$       & $45.6\ \ $  \\
multi-stage \cite{qian2015fine}       &  $B$         & $B$       & $70.31$   \\
Simple tech \cite{xie2016simple}      &  $B$         & $B$       & $43.15 $   \\
FOAF \cite{zhang2016fused}            &  $B$         & $B$       & $74.49 $   \\
PBC \cite{huang2016pbc}               &  $B$         & $B$       & $78.3\ \ $  \\
\hline
Attention \cite{sermanet2014attention}&  $-$         & $-$       & $76.8\ \ $  \\
Constellation \cite{simon2015neural}  &  $-$         & $-$       & $68.61 $  \\
Picking deep \cite{zhang2016picking}  &  $-$         & $-$       & $71.96   $  \\
FOAF \cite{zhang2016fused}            &  $-$         & $-$       & $68.66 $   \\
W-supervised \cite{weakly}            &  $-$         & $-$       & $80.43 $   \\
Friend-or-Foe \cite{xu2017friend}     &  $-$         & $-$       & $71.4\ \ $  \\
GoogLeNet-ft                          &  $-$         & $-$       & $82.6\ \ $  \\
Ours-GoogLeNet-based                  &  $-$         & $-$       & $83.6\ \ $   \\
VGG-19-ft                             &  $-$         & $-$       & $80.97$   \\
Ours-VGG-19-based                     &  $-$         & $-$       & $83.0\ \ $   \\
VGG-CNN-S-ft                          &  $-$         & $-$       & $72.90$   \\
Ours-VGG-CNN-S-based                  &  $-$         & $-$       & $73.7\ \ $   \\
\hline
\end{tabular}
}
\end{center}
\caption{Comparison of different classification methods on Stanford dogs dataset~\cite{khosla2011novel}. `Anno' stands for `annotations', `B' for `bounding box', `P' for `part location' and `ft' for `fine-tune'. The rounding precision of other methods may differ a little in other citations.}
\label{tab:dogs result}
\end{table}

\section{Conclusion}
\label{sec:Conclusion}
In this paper, we have proposed a novel CNN-based fine-grained categorization method for the general problem setting that only class labels are available during training with no other annotations.  Not only does it achieve accuracy comparable to the state-of-the-art, but it also sheds some light on ingenious use of the features learned by CNN which can find wide applications well beyond the current fine-grained categorization task.  Instead of trying to deepen the network architecture as many other researchers did, we seek to exploit the hidden layer feature maps learned to achieve robust object detection, part detection, and pose estimation which together can boost the classification accuracy significantly.  Our future work will consider extending this approach to other tasks.

\clearpage

\section{Acknowledgements }
This work took me long time to complete and my supervisor Prof. D.Y. Yeung and senior labmate Lin Sun have offered me grate guidance and support all the way. Many thanks to them.

\section{ Funding sources }
I am fully supported by the Hong Kong University of Science and Technology.

\section*{References}

\bibliography{stbibfile}

\end{document}